\relax
\documentclass[letterpaper]{article} 

\usepackage{ragged2e}
\usepackage{times}  
\usepackage{helvet} 
\usepackage{courier}  
\usepackage[hyphens]{url}  
\usepackage{graphicx} 
\usepackage{natbib}  
\usepackage{caption} 
\usepackage{comment}
\usepackage{amsmath,amssymb} 
\usepackage{color}
\usepackage{algorithm}
\usepackage{algorithmic}
\usepackage{microtype}
\usepackage{subfig}
\usepackage{booktabs}
\usepackage{amsmath,amssymb,amsfonts}
\usepackage{mathtools}
\usepackage{soul}
\usepackage{hyperref}
\usepackage{cleveref}
\usepackage{xspace}
\usepackage{fullpage}

\title{Perceptually Constrained Adversarial Attacks \\
}
\author{

Muhammad Zaid Hameed$^1$,  Andr\'as Gy\"orgy$^{2}$\\

$^1$Department of Computing,
Imperial College London, UK \\
$^2$DeepMind, London, UK \\
Email: muhammad.hameed13@imperial.ac.uk, agyorgy@google.com
}

\date{}

\newcommand{\xent}{\mathrm{xent}}
\newcommand{\CW}{\mathrm{CW}}
\newcommand{\Madry}{\mathrm{Madry}}
\newcommand{\Madryx}{\mathrm{Madry_{xent}}}
\newcommand{\Madryc}{\mathrm{Madry_{CW}}}
\newcommand{\Free}{\mathrm{Free}}
\newcommand{\Freex}{\mathrm{Free_{xent}}}
\newcommand{\Freec}{\mathrm{Free_{CW}}}
\newcommand{\Feature}{\mathrm{Feature}}
\newcommand{\Featurex}{\mathrm{Feature_{xent}}}
\newcommand{\Featurec}{\mathrm{Feature_{CW}}}
\newcommand{\PGD}{\mathrm{PGD}}
\newcommand{\ENET}{\mathrm{ENet}}
\newcommand{\SSIM}{\mathrm{SSIM}}
\newcommand{\SSIME}{\mathrm{SSIM_{E}}}
\newcommand{\R}{\mathbb{R}}
\DeclareMathOperator*{\argmax}{arg\,max}
\DeclareMathOperator*{\argmin}{arg\,min}

\renewcommand{\S}{\mathcal{S}}

\newcommand{\K}{\mathcal{K}}
\newcommand{\blambda}{\boldsymbol{\lambda}}

\renewcommand{\S}{\mathcal{S}}

\newif\ifmnist
\mnistfalse
\mnisttrue
\newif\ifcifar
\cifarfalse
\newif\ifssimscale
\ssimscaletrue

\begin{document}

\maketitle

\begin{abstract}
Motivated by previous observations that the usually applied $L_p$ norms ($p=1,2,\infty$) do not capture the perceptual quality of adversarial examples in image classification, we propose to replace these norms with the structural similarity index (SSIM) measure, which was developed originally to measure the perceptual similarity of images. Through extensive experiments with adversarially trained classifiers for MNIST and CIFAR-10, we demonstrate that our SSIM-constrained adversarial attacks can break state-of-the-art adversarially trained classifiers and achieve similar or larger success rate than the elastic net attack, while consistently providing adversarial images of better perceptual quality. Utilizing SSIM to automatically identify and disallow adversarial images of low quality, we evaluate the performance of several defense schemes in a perceptually much more meaningful way than was done previously in the literature.

\end{abstract}

\section{Introduction}\label{sec: introduction}

In recent years, the advances in machine learning have enabled solving problems in artificial intelligence, such as understanding speech, natural languages, or images, at an unprecedented accuracy, enabling such systems being deployed in practical applications. This growing use of machine learning methods has given rise to concerns about their security and reliability, especially for one of the most practical methods, known as deep learning (DL). In particular, it has been demonstrated that modifications to their input (e.g., an image or speech signal), imperceptible to humans, can fool deep learning models and make them commit unexpected errors. These modifications are known as adversarial attacks or adversarial input perturbations \citep{goodfellow2014explaining,kurakin2016adversarial,chen2017ead,carlini2017towards}. 

Specifically, in image classification the goal of adversarial attacks is to fool a classifier such that the created adversarial images have similar high-level features as the original inputs, so that for a human oracle they belong to the same class (a stricter version of this requirement is that the difference of the original and the adversarial images should be imperceptible for a human oracle). On the other hand, in practice, in most cases the quality of the adversarial perturbations is measured through pixelwise distortion measures in some $L_{p}$ distance (with $p \in \{1, 2, \infty \}$), operationalizing the assumption that if two images have small distance in these distortion measures than they are perceptually similar, that is, they have similar high-level features. At the same time, \citet{sharif2018suitability} demonstrated that having a small $L_p$ distance is both unnecessary and insufficient for perceptual similarity and proposed the use of other similarity measures available in the literature \citep{wang2004image,wang2002universal,yee2001spatiotemporal} which are more aligned with human perception. \citet{sharif2018suitability} also showed that these perception-based similarity measures are not fully satisfactory either, for example, they cannot handle geometric transformations such as rotations or translations, and carefully introducing patches from other images might produce significantly different values for perceptual similarity measures. 

\begin{figure}
\centering
\vspace{-0.9cm}
    \captionsetup[subfloat]{format=hang,singlelinecheck=false,justification=centering,size=small}
      \subfloat[Original \protect\\
      \small label: 3
      ]{\includegraphics[width=0.2\textwidth]{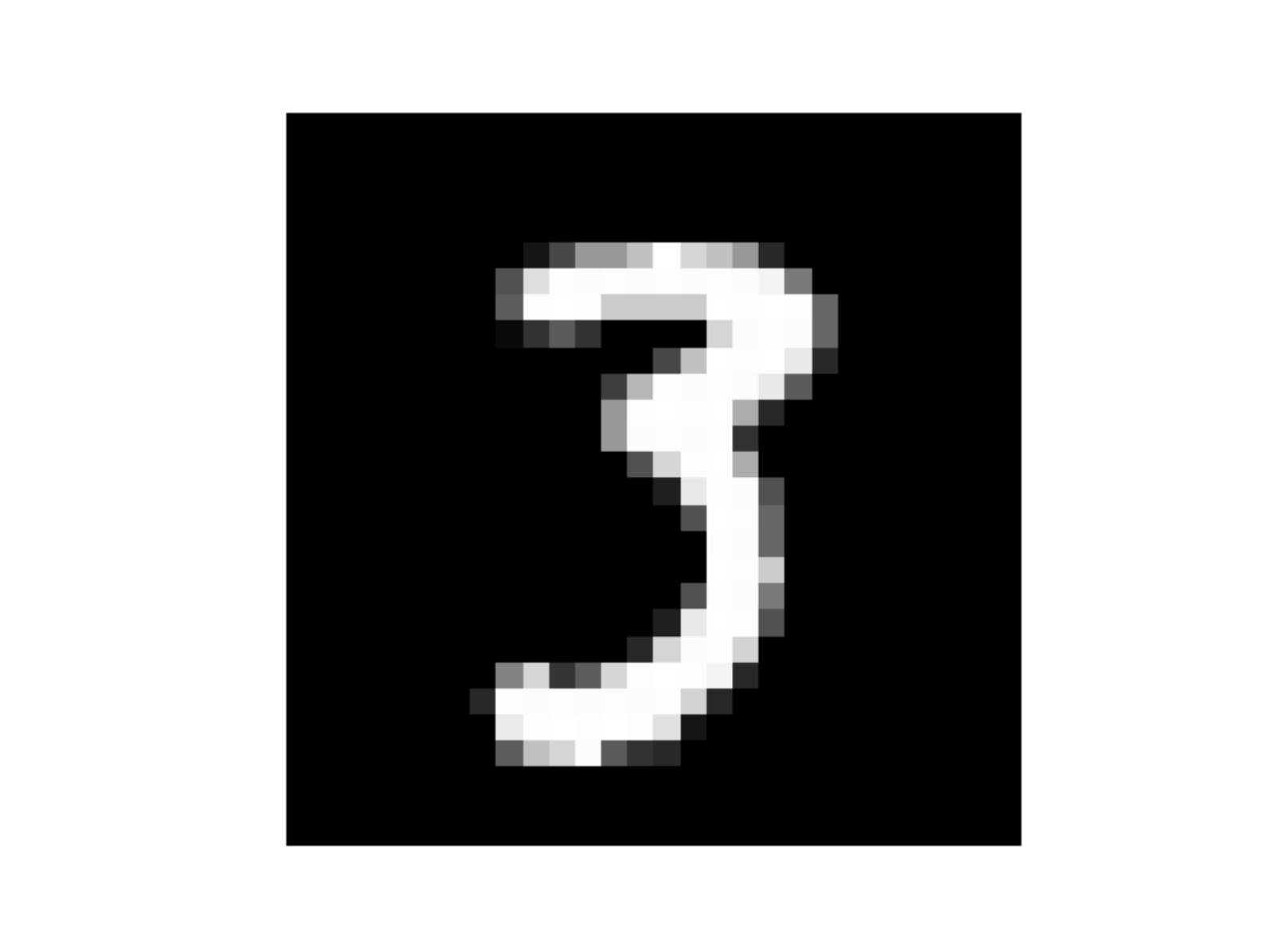}}
      \subfloat[$\ENET$ \protect\\  
      \small label: 9\protect\\ 
      \small $L_{1}$: 10.95, 
      \small $L_{2}$: 2.69  \protect\\ 
      \small $L_{\infty}$: 1.0, 
      \small SSIM: 0.39
      ]{\includegraphics[width=0.2\textwidth]{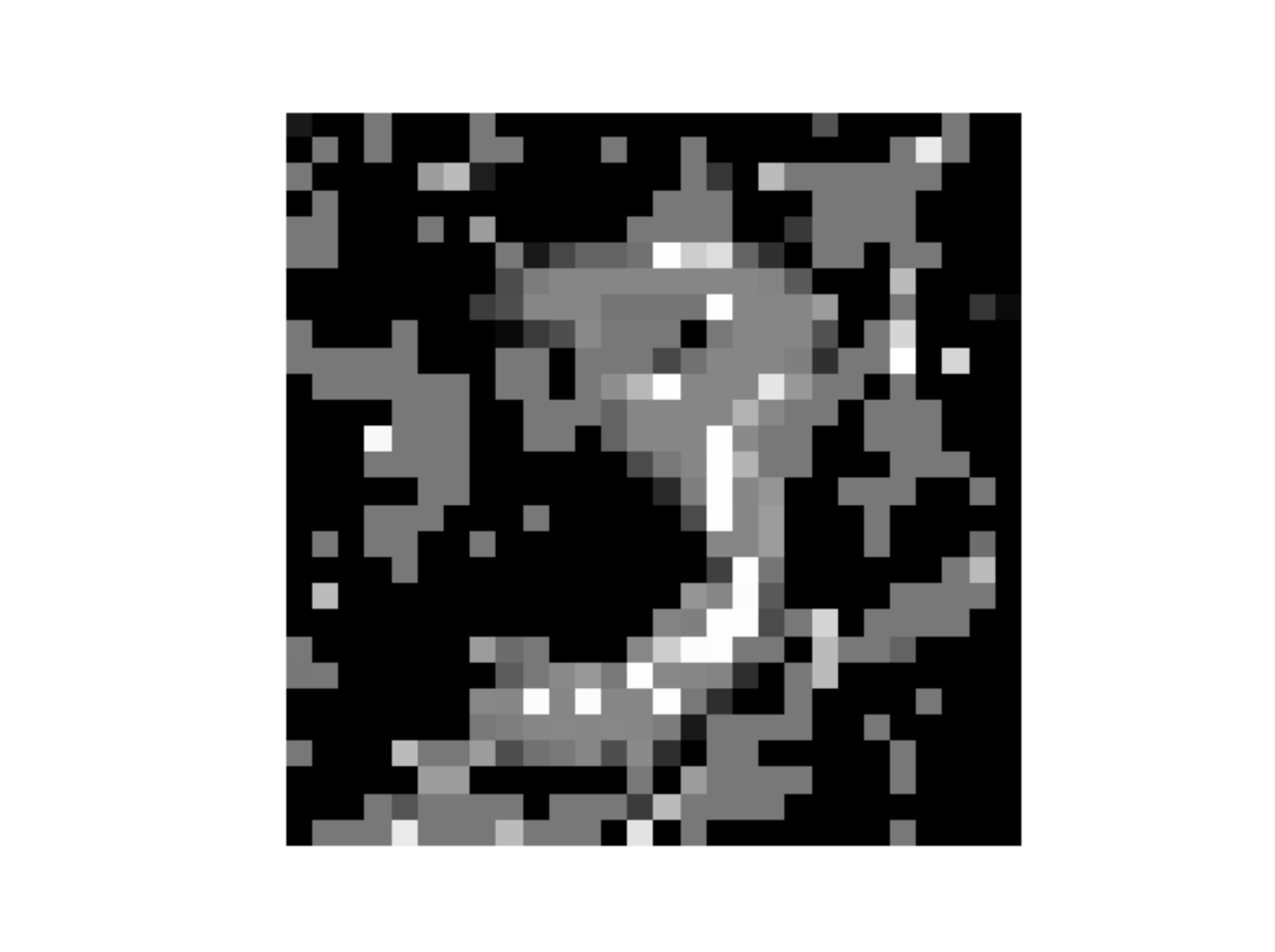}}
      \subfloat[$\SSIME$ \protect\\ 
      \small label: 8\protect\\
      \small $L_{1}$: 10.88, 
      \small $L_{2}$: 2.67 \protect\\
      \small $L_{\infty}$: 0.99, 
      \small SSIM = 0.58
      ]{\includegraphics[width=0.2\textwidth]{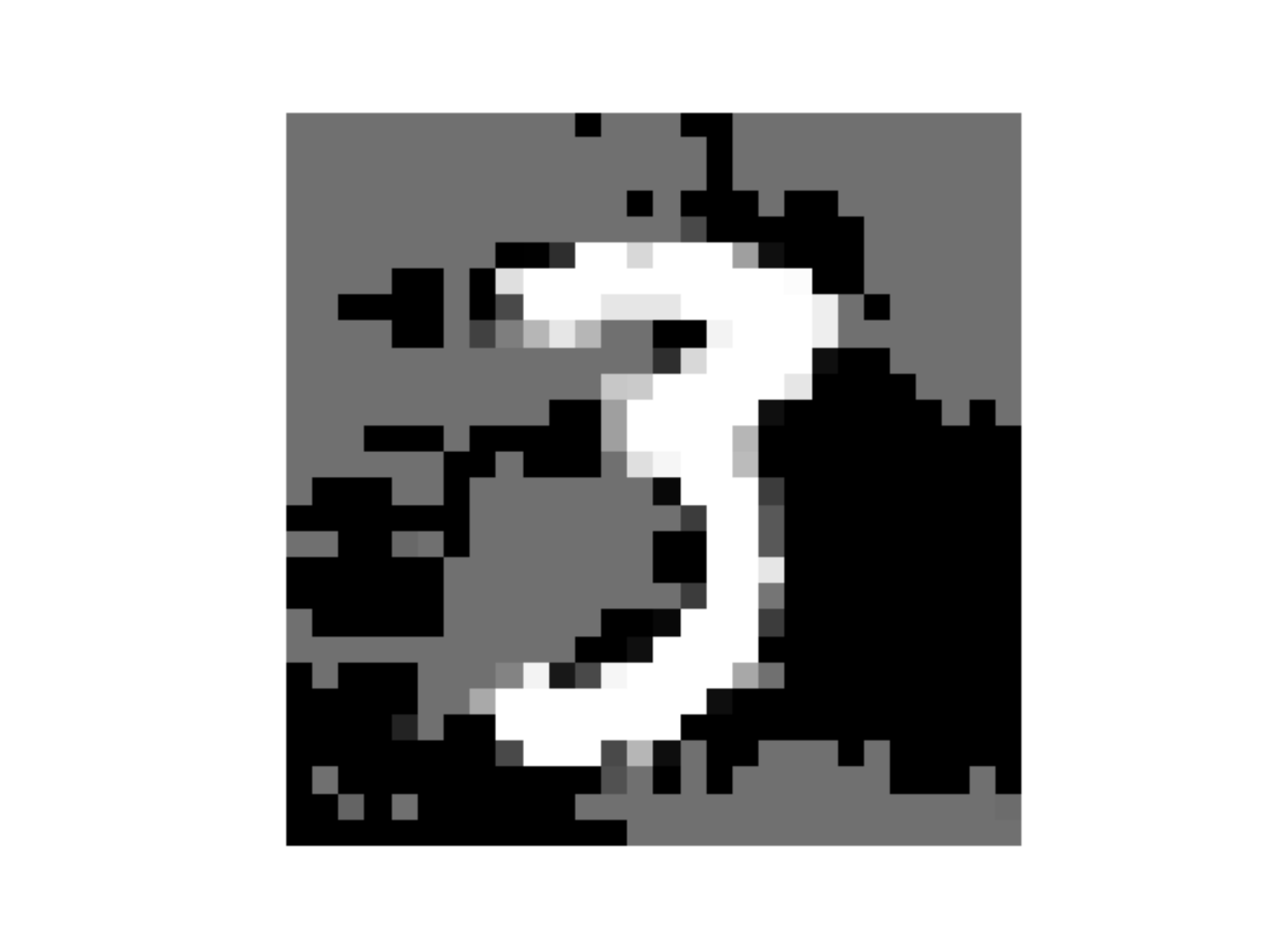}}
	\vspace{-0.2cm}
      \caption{$L_{p}$ distances do not convey significant information about perceptual similarity between images. Subfigures (a), (b), (c), respectively, show an original MNIST digit, an adversarial version created by the elastic net ($\ENET$) attack and our SSIM-based $\SSIME$ attack. Class labels given by the classifier, $L_p$ distances for $p\in\{1,2,\infty\}$, and SSIM are also shown.
      }
      \label{fig: need_for_ssim}
\end{figure}

A direct implication of this result is that the adversarial attacks using pixelwise $L_{p}$-distances as a proxy for perceptual similarity may produce adversarial images which will be near in $L_{p}$-distance to the original images, but either belong to some other class or are destroyed beyond recognition. \Cref{fig: need_for_ssim} shows such a case when two adversarial attacks produce successful adversarial images with almost equal $L_{p}$-distances from the original image  (simultaneously for $p=1,2,\infty$), but one of the images (obtained by the state-of-the-art elastic net attack of \citealt{chen2017ead}) is nearly destroyed beyond recognition, while the other adversarial image (created by one of our proposed attacks based on perceptual similarity) could be easily classified by a human oracle. At the same time, defense strategies which use the same underlying assumption that images close in some $L_{p}$-distance belong to the same class suffer from the fact that they try to assign the same class label to different images with small $L_{p}$-distances, instead of being perceptually similar. Thus, adversarial images which are not near in some $L_{p}$-distance but share the same high-level features can potentially break these defense methods.

In order to alleviate these shortcomings of current adversarial attack and defense approaches, we explore perceptual similarity in determining the quality of adversarial attacks and the robustness of defense strategies against these attacks. More specifically, our contributions in this work are as follows:
\begin{itemize}
    \item We propose to use the structural similarity index (SSIM) measure \citep{wang2004image} as a quality metric for perceptual similarity of adversarial images and show that SSIM is quite effective in quickly and automatically finding adversarial images of low quality, for example, when an attack destroys the whole image. 
    \item We propose SSIM-based adversarial attacks, and show that they are able to break state-of-the-art defense schemes, achieving similar or larger success rate than the state-of-the-art elastic net attack ($\ENET$) \citep{chen2017ead} (and also much stronger than the simple $L_\infty$-constrained projected gradient descent, $\PGD$, attack \citep{madry2017towards}), while consistently producing adversarial images of better perceptual quality. 
    
\end{itemize}

\subsection{Related Work}\label{sec: related_work}

Adversarial examples were first discovered by \citet{szegedy2013intriguing}, who found that for state-of-the-art CNN-based image classifiers, for almost all images it is possible to create a perturbation with a small $L_{2}$ norm such that when added to the input image, it changes the class label predicted by the classifier while the resulting input image is visually similar to the original input image. Since then a multitude of schemes were suggested to create such adversarial perturbations, mostly for classification; the most popular class of these methods phrases the task as a constrained optimization problem in the input space, where the error (and often the confidence in the prediction) of the classifier is maximized over the set of inputs which are close to the original input in some $L_p$ distances
 \citep{goodfellow2014explaining,kurakin2016adversarial,madry2017towards,papernot2017practical,tramer2017ensemble,carlini2017towards,chen2017ead,rony2019decoupling}.

In parallel, several defense methods were developed to train classifiers which are more robust to such adversarial perturbations. Most of these schemes use some form of adversarial training, which uses carefully created adversarially examples during the training of a classifier \cite{huang2015,madry2017towards,yan2018deep,gowal2018effectiveness,shafahi2019adversarial,rony2019decoupling}. Following the same principles, \citet{feature_scatter} introduced perturbations by constraining the changes in the feature space instead of the input space, further increasing the performance of these attacks.\footnote{These methods can be combined with other approaches, such as using ensembles for classifiers \citep{pmlr-v97-cohen19c} or using unlabelled data in the training \citep{unlabeled_data_2019}, but looking at such combinations is orthogonal to the questions we consider here and, hence, are out of the scope of the paper.}
Yet, this defense scheme can also be broken, e.g., with the elastic net attack \citep{chen2017ead}, however, in such successful attacks the images are often heavily distorted and sometimes even destroyed beyond recognition.

In this work we address this problem by constraining adversarial perturbations using perceptual similarity measures, namely SSIM \citep{wang2004image}, and show that even the best defense schemes (to date) can be (almost) completely broken by visually acceptable adversarial images.
Two recent papers \citep{gragnaniello2019perceptual,zhao2019towards}, developed in parallel of our work (one of them only published on arXiv), also use perceptual distance measures: \citet{gragnaniello2019perceptual} proposed to use SSIM to create adversarial examples, while \citet{zhao2019towards} used the perceptual color distance for the same purpose. Both works demonstrate that their proposed attack method works well against standard classifiers with simple defense schemes, such as JPEG compression \citep{dziugaite2016study}, bit-depth reduction \citep{xu2017feature,guo2018countering}, or nearest neighbor classification. Since such classifiers are easily broken even by $L_p$-constrained attacks (such as $\PGD$ by \citealt{madry2017towards} or $\ENET$ by \citealt{chen2017ead}) while introducing little visual artifacts, the evaluation of these attacks is limited. In contrast, we study the performance of our attack methods against adversarially trained networks, which is a much more challenging task \citep{athalye2018obfuscated}, and show that the attacks are able to break even state-of-the-art defense schemes while producing high-quality adversarial images.

\section{Perceptually Constrained Adversarial Attacks }\label{sec: unrestricted_adversarial_examples}

We consider a score-based classifier of inputs $x \in \R^n$ to a set of labels $\S =\{1, 2, \ldots, S\}$, defined by the score function $f: \R^n \times \S \to \R$, assigning a label  $\hat{s} \in \argmax_{s \in \S} f(x,s)$ to $x$. With a slight abuse of notation, throughout we also use $f$ to denote the classifier, $f(x)$ to denote the label assigned to $x$, and $f_s(x)=f(x,s)$ to denote the score of class $s$.

Let $x$ be an image correctly classified by $f$, that is, $s = f(x)$, the true label. An adversarial attack on this classifier $f$ aims to modifiy the input image $x$ with a perturbation $\delta \in \R^{n}$, such that $f(x + \delta) \neq f(x)$, that is, the class label is changed from the original correct prediction.    
Since $x$ is correctly classified, $f_s(x) - \max_{s' \neq s}(f_{s'}(x)) > 0$, and $\delta$ is a successful adversarial perturbation if the class label changes, that is, 
$f_s(x + \delta) - \max_{s' \neq s}(f_{s'}(x + \delta))< 0$. To find such a $\delta$, following  \citet{carlini2017towards}, we minimize a loss function $\ell(x, \delta)$ defined
such that $\ell(x, \delta) = 0$ only when this condition is met:
\begin{equation}
\label{eq: car_loss}
\ell(x, \delta) = c \left(f_s(x + \delta) - \max_{s' \neq s}f_{s'}(x + \delta) + \K\right)^+,
\end{equation}
where $\K$ represents some required margin, $c>0$ is a scaling factor (the role of $c$ will be to balance the loss function and the additional regularization terms in Eq.~\ref{eq: opti})
and $a^+$ is a short-hand for $\max(a, 0)$ (note that $\ell(x, \delta)$ depends on $x$ and $\delta$ through their sum $x + \delta$ and the true label $s$ of $x$). 
A solution of the resulting unconstrained optimization problem $\argmin_{\delta}\ell(x, \delta)$ is an adversarial example with a score-margin of at least $\K$ when $\ell(x,\delta)=0$.
To keep the resulting image $x+\delta$ close to $x$, the attack methods in the literature use a constrained formulation requiring that some similarity measure $Q(x, x+ \delta)$ between the input $x$ and its perturbed version $x + \delta$ is lower bounded by some value $\zeta$, leading to the optimization problem
\begin{equation}\label{eq: opti}
\argmin_{\delta} \ell(x, \delta) \text{ such that } Q(x, x + \delta) \ge \zeta.
\end{equation}
In the literature, $Q$ is some negative $L_p$ distance, that is, $Q(x,x+\delta)=-\|\delta\|_p$ with $p \in \{1, 2, \infty\}$ usually,
and instead of the constrained optimization formalization \eqref{eq: opti}, a Lagrangian formulation
\begin{equation}
   \argmin_\delta  \ell(x, \delta) + \lambda (\zeta - Q(x, x+ \delta)) 
   \label{eq: lagrn}
\end{equation}
is used with some $\lambda>0$ \citep{goodfellow2014explaining,kurakin2016adversarial,chen2017ead,carlini2017towards}.

As the similarity measure $Q$, we consider the structural similarity index (SSIM) measure  \citep{wang2004image} defined, for gray-scale images $x, y \in \R^n$, as
\begin{equation}
\label{eq:SSIM_measure}
    {\SSIM(x, y) = [l(x, y)]^{\alpha}\cdot [c(x, y)]^{\xi} \cdot [s(x, y)]^{\gamma}},
\end{equation}
where $l(x, y)  = \frac{2\mu_x\mu_y + C_1}{\mu_x^2 +\mu_y^2 + C_1}$ is luminance with $\mu_{x}$ and $\mu_y$ denoting the mean pixel value of $x$ and $y$, resp., 
$c(x, y)  = \frac{2\sigma_{x}\sigma_{y} + C_{2}}{\sigma_{x}^2 +\sigma_{y}^2 + C_2}$ is the contrast function with $\sigma_{x}$ and $\sigma_y$ denoting the standard deviation of  pixel values of $x$ and $y$, resp., and $s(x, y)  = \frac{\rm{cov}_{xy} + C_{3}}{\sigma_{x}\sigma_{y} + C_3}$ is the structure comparison function with $\rm{cov}_{xy}$ denoting the pixelwise covariance between images $x$ and $y$; here $C_1, C_2, C_3 > 0$ and $\alpha, \xi, \gamma>0$ are appropriate constants.
In case of RGB images, the SSIM is calculated for each channel and then the average SSIM value across the different color channels is computed. 
For $\alpha=\xi=\gamma=1$ and $C_{3}=\frac{C_{2}}{2}$, \eqref{eq:SSIM_measure} simplifies to
\begin{equation}\label{eq: ssim_combined}
\SSIM(x, y) =\underbrace{\frac{2 \mu_{x}\mu_{y} + C_1}{\mu_{x}^2 +\mu_{y}^2 + C_1}}_{S_1}\cdot\underbrace{\frac{ 2 \ \rm{cov}_{x y} + C_2}{\sigma_{x}^2 +\sigma_{y}^2 + C_2}}_{S_2}~.
\end{equation}
We use the same values for all parameters ($C_{1}, C_{2}, C_{3}$) as \citet{wang2004image} (specifically, $\sqrt{C_1}$ and $\sqrt{C_2}$ are 0.01, resp., 0.03 times the range of the coordinates of $x$).
To compute the adversarial attack, we would like to solve the optimization problem \eqref{eq: lagrn} with $Q=\SSIM$. The non-cocavity of $\SSIM$ makes this optimization problem hard (even though $f$ is non-convex by itself, this is an issue in practice). Therefore, following \citet{brunet2011mathematical}, we aim to optimize separately the two terms $S_1=S_1(\mu_x,\mu_y)$ and $S_2=S_2(x-\mu_x,y-\mu_y)$ in \eqref{eq: ssim_combined}, where, with a slight abuse of notation, $x-\mu_x$ denotes an $n$-dimensional vector with its $i$th coordinate being defined as $x_i-\mu_x$ (and $y-\mu_y$ is defined similarly). 
The benefit of this approach is that $-S_1$ and $-S_2$ can be shown to be quasi-convex (i.e., all their level sets are convex) \citep{brunet2011mathematical}: Consider the normalized mean squared error (NMSE) function $\mathrm{NMSE}(u, v, C) = \frac{\|u - v\|^{2}}{\|u\|^{2} + \|v\|^{2} + C}$ 
for $u,v \in \R^d$ and $C\ge 0$ which is quasi-convex in $u$ on the set $\{u: \langle u,v \rangle \ge -C/2\}$ for a fixed $v$. Then we have
\begin{align*}
S_{1}(\mu_{x}, \mu_{y}) & = 1 - \mathrm{NMSE}(\mu_{x}, \mu_{y}, C_{1}), \\
S_{2}(x - \mu_{x}, y - \mu_{y}) &= 1 - \mathrm{NMSE}(x - \mu_{x}, y - \mu_{y}, C_{2}),
\end{align*}
showing quasi-convexity for the appropriate regions for $x$ for a fixed $y$: $-S_1$ is quasi convex on 
 $P_{1} = \{ x:  \mu_{x} \mu_{y} \geq -\frac{C_{1}}{2} \}$ and $-S_2$ is quasi-convex on 
$P_{2} = \{ x - \mu_{x}: \ \langle x - \mu_{x}, y - \mu_{y} \rangle \geq -\frac{C_{2}}{2} \}$. 

Now, to simplify the minimization process, instead of the constraint $\SSIM(x,y) \ge \zeta$, we consider $S_1 \ge \zeta_1$ on $P_1$ and $S_2 \ge \zeta_2$ on $P_2$. Ensuring 
the lower bounds on $S_1$ and $S_2$ and that the adversarial image falls into $P_1 \cap P_2$, we have the constraints
\begin{align*}
g_1(y,x) &=  \zeta_{1} - \left(1 - \tfrac{( \mu_{x} - \mu_{y} )^{2}}{\mu_{x}^{2} + \mu_{y}^{2} + C_{1}}\right) \le 0; \\
g_2(y,x) &= \zeta_{2} - \left(1 - \tfrac{\| (x -\mu_{x}) - (y -\mu_{y}) \|^{2}}{\|(x -\mu_{x})\|^{2} + \|y -\mu_{y}\|^{2} + C_{2}} \right) \le 0;  \\
g_3(y,x) &= -2 \mu_{x} \mu_{y} - C_{1} \le 0; \\
g_4(y,x) & = -2 \langle x - \mu_{x}, y - \mu_{y} \rangle -C_{2} \le 0.
\end{align*}
Note that for $[0,1]$-valued pixels, $g_3(y,x) \le 0$ always holds with our choice of $C_1$.
To minimize $\ell(x,\delta)$ subject to the constraints above, we formulate the Lagrangian
\begin{align}
L(\delta, \blambda) &=  \ell(x, \delta) + \sum_{i=1}^{4}\lambda_{i} g_{i}(x + \delta, x),
   \label{eq: lagrn_mod}
\end{align}
which is a relaxed version of \eqref{eq: lagrn} (here $\blambda=(\lambda_1,\ldots,\lambda_4)$), and 
simultaneously minimize it over $\delta$ and maximize it in the Lagrangian multipliers $\blambda$, similarly to \citet{cotter2019two}. The resulting gradient-based (first-order) optimization method is presented in \Cref{algo: lagopt}, where to make an update in $\delta$ and $\blambda$, we use first-order optimization algorithms $A_\delta$ and $A_{\blambda}$, where $A_i(x,\nabla_x)$ provides the next iteration from $x$ using gradient $\nabla_x$. In practice, these algorithm can be chosen, e.g., as gradient descent or Adam \citep{kingma2014adam} (e.g., when $A_\delta$ is the gradient descent algorithm, 
$A_\delta(\delta^{t},  \nabla_{\delta} L(\delta^{t}, \blambda^{t}))= \delta^t - \eta \nabla_{\delta} L(\delta^{t}, \blambda^{t})$ for some step size $\eta>0$).  Of course, since our optimization problem is non-convex, \Cref{algo: lagopt} is not guaranteed to find the solution of the constrained  minimization problem.  

\begin{algorithm}[tb]
   \caption{Optimization of the Lagrangian \eqref{eq: lagrn_mod}.} 
   \label{algo: lagopt}
\begin{algorithmic}
\STATE{\bfseries Input:} $T \in \mathbb{N}$, iterative gradient-based optimization algorithms $A_\delta$ and $A_{\blambda}$
   \STATE {\bfseries Initialize:} $\delta^{1}$ = 0, $\blambda^{1}$ = 0
   \FOR{$t = 1,\ldots,T$}
  \STATE Update $\hat{\delta}^{t+1} = A_\delta(\delta^{t},  \nabla_{\delta} L(\delta^{t}, \blambda^{t}))$
  \STATE Update $\delta^{t+1} = \mathrm{clip}(x + \hat{\delta}^{t+1}, [0, 1]^n) - x$
  \STATE Update $\blambda^{t+1} = \max\{A_{\blambda}(\blambda^{t},-\nabla_{\blambda} L(\delta^{t}, \blambda^{t})), 0\}$
  \ENDFOR
  \STATE  Return $x + \delta^{T}$
\end{algorithmic}
\end{algorithm}

While in \Cref{algo: lagopt} we initialize $\delta$ and $\blambda$ with zero, in practice it is often beneficial to start from a reasonably good local optimum. For this reason, we also consider starting the optimization from the perturbation given by the elastic net ($\ENET$) attack \citep{chen2017ead}, obtained as the (approximate) solution of the loss minimization with an elastic loss penalty:
\begin{align}\label{eq: elastic_loss}
\delta^*  &= \argmin_{x+\delta \in [0, 1]^n} \ell(x, \delta) + \beta \|\delta\|_1 +  \|\delta\|_2^2
\end{align}
for some $\beta>0$. 
Note that this is a special case of \eqref{eq: lagrn}  with $Q(x,x+\delta) = - \beta \|\delta\|_1 -  \|\delta\|_2^2$. 
Our $\ENET$-initialized attack can be interpreted as an SSIM-based perceptual improvement over the $\ENET$ attack. In the $\ENET$ attack,  \citet{chen2017ead} proposed to optimize the elastic-loss objective \eqref{eq: elastic_loss} by some first-order optimization algorithm, augmented with a binary search over the parameter $c$, used in the definition of the loss function (see \Cref{algo: binsearch} in the appendix)
, which we also adopt in computing our attacks.

\section{Experimental Evaluation}\label{sec: experimental_evaluation}

In this section we present experiments showing the effectiveness of our SSIM-based attacks. Unlike concurrent work \citep{gragnaniello2019perceptual,zhao2019towards}, we consider adversarially trained networks for both the MNIST \citep{MNIST} and the CIFAR-10 \citep{CIFAR10} datasets.
We compare our attack methods to standard and state-of-the-art adversarial attacks: the elastic net attack ($\ENET$) \citep{chen2017ead} given in \eqref{eq: elastic_loss}, which aims to obtain adversarial images with perturbations of small $L_1$- and $L_2$-norms, and the projected gradient descent attack ($\PGD$) in $L_{\infty}$-norm \citep{madry2017towards}. 

The basic version of the $\SSIM$ attack (initialized at zero) is denoted by $\SSIM$, while the one which is initialized at the $\ENET$ attack is denoted by $\SSIME$. In the implementation (\Cref{algo: lagopt}), we used gradient descent as $A_\delta$ and
the Adam optimizer \citep{kingma2014adam} as $A_{\blambda}$, as this combination led to the most successful adversarial perturbations with large SSIM values. To properly tune the scaling coefficient $c$ in \eqref{eq: car_loss}, we use binary search, as proposed by \citep{chen2017ead} for the $\ENET$ attack (details are given in \Cref{algo: binsearch} in the appendix); the same method is used in $\ENET$. In $\SSIME$, we use the same $c$ in the optimization as the one obtained by $\ENET$ during the initialization.

In the experiments we set the confidence $\K$ to 0 in the loss function \eqref{eq: car_loss}.
$\ENET$ was implemented with $\beta=0.01$. We used 9 binary search steps, each of which involved 1000 iterations each with initial learning rate $0.01$ reduced with the square-root of the number of iterations. 

For MNIST we analyze the performance of these attacks for a single classifier, while for CIFAR-10 we consider three defense schemes.

\subsection{SSIM Attack on MNIST}

\begin{figure}
    \centering
    \vspace{-0.8cm}
    \includegraphics[width=0.45\columnwidth, trim={5inch 13inch 4inch 16inch},clip]{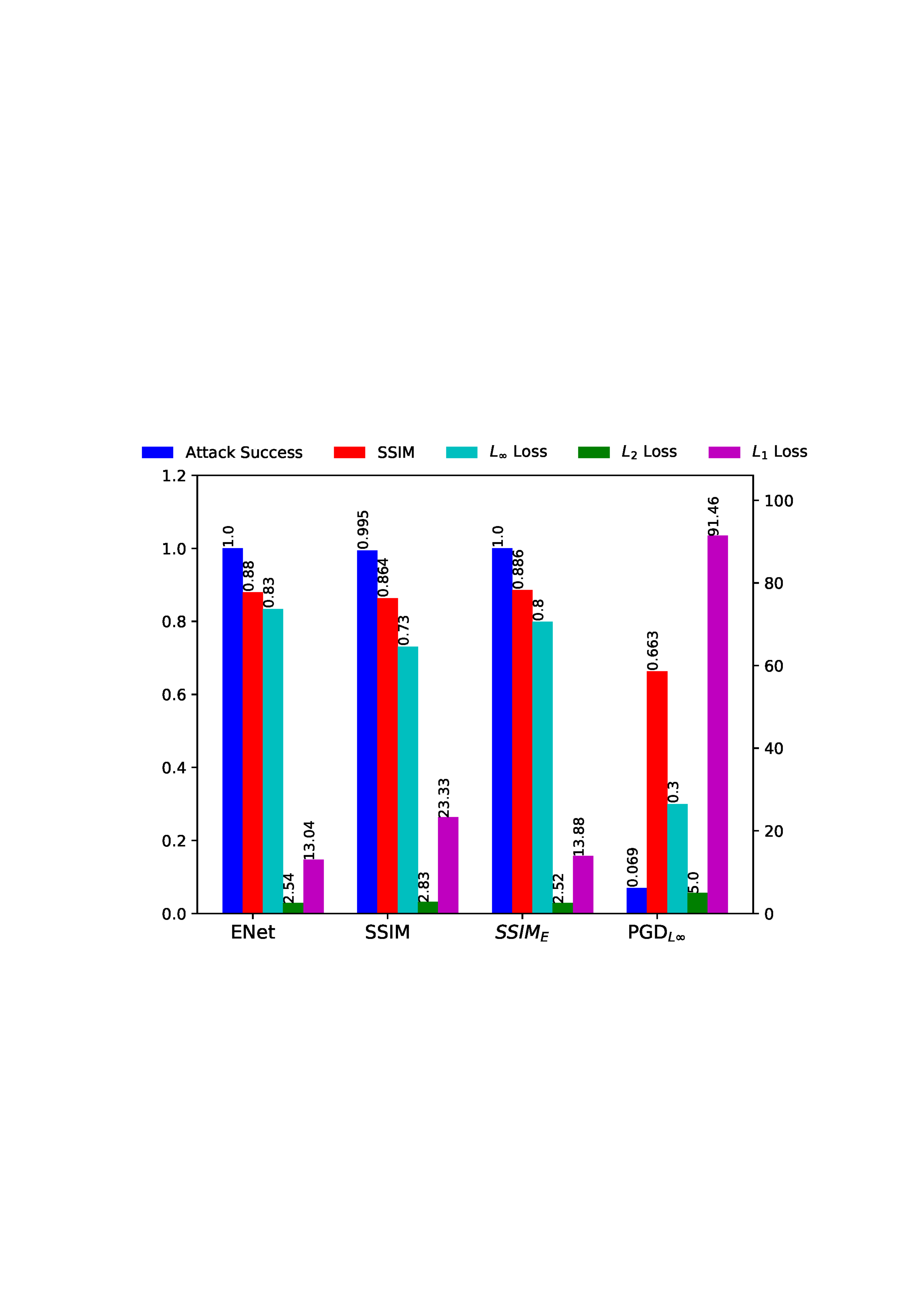}
    \caption{Performance of adversarial attacks on MNIST with adversarial training.}
    \label{fig: mnist_adv_performance}
\end{figure}

For the MNIST dataset (with pixel intensities scaled to $[0,1]$),  we compare the performance of the attack methods against an adversarially trained network. We use a convolutional neural network from the Cleverhans library \citep{papernot2018cleverhans} with three convolution layers of 64, 128 and 256 filters (of size 3x3), respectively, with ReLU activations and one fully connected layer of size 128.  Following \citet{madry2017towards}, the network was trained with a mix of clean and adversarial images generated by 20 iterations of the $\PGD$ attack with maximum $L_\infty$-perturbation $\epsilon=0.3$. \Cref{fig: mnist_adv_performance} shows the performance of the adversarial attacks used for comparison in terms of attack success, average SSIM value and $L_{p}$ distortions for $p\in\{1, 2, \infty\}$ for successful adversarial images.

\begin{figure}[t]
\centering
\begin{tabular}{cc}
    \includegraphics[height = 6cm, trim={0.3cm 0cm 2cm 1.5cm},clip]{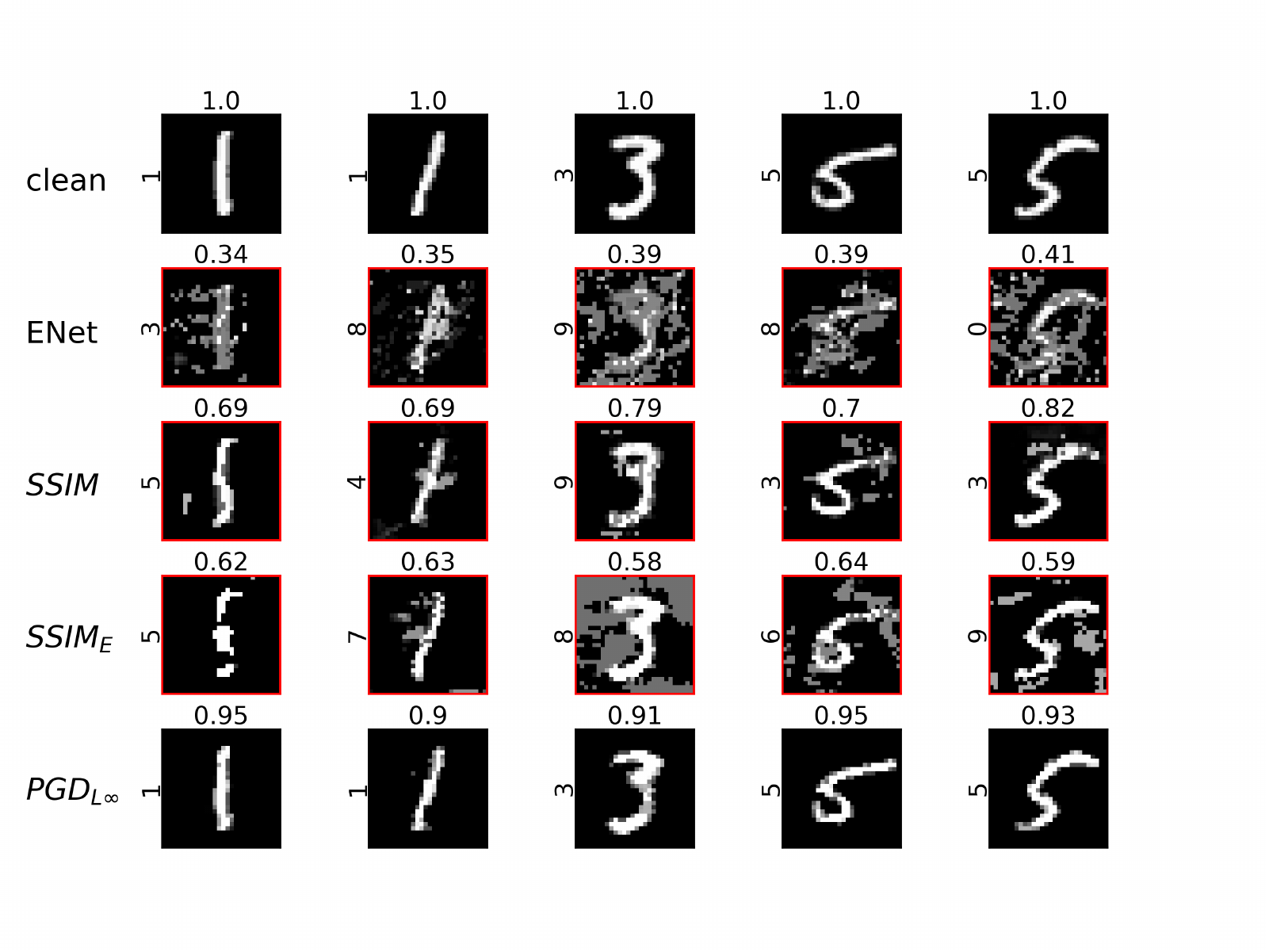}
    &  \includegraphics[height = 6cm, trim={0.3cm 0cm 2cm 1.5cm}, clip]{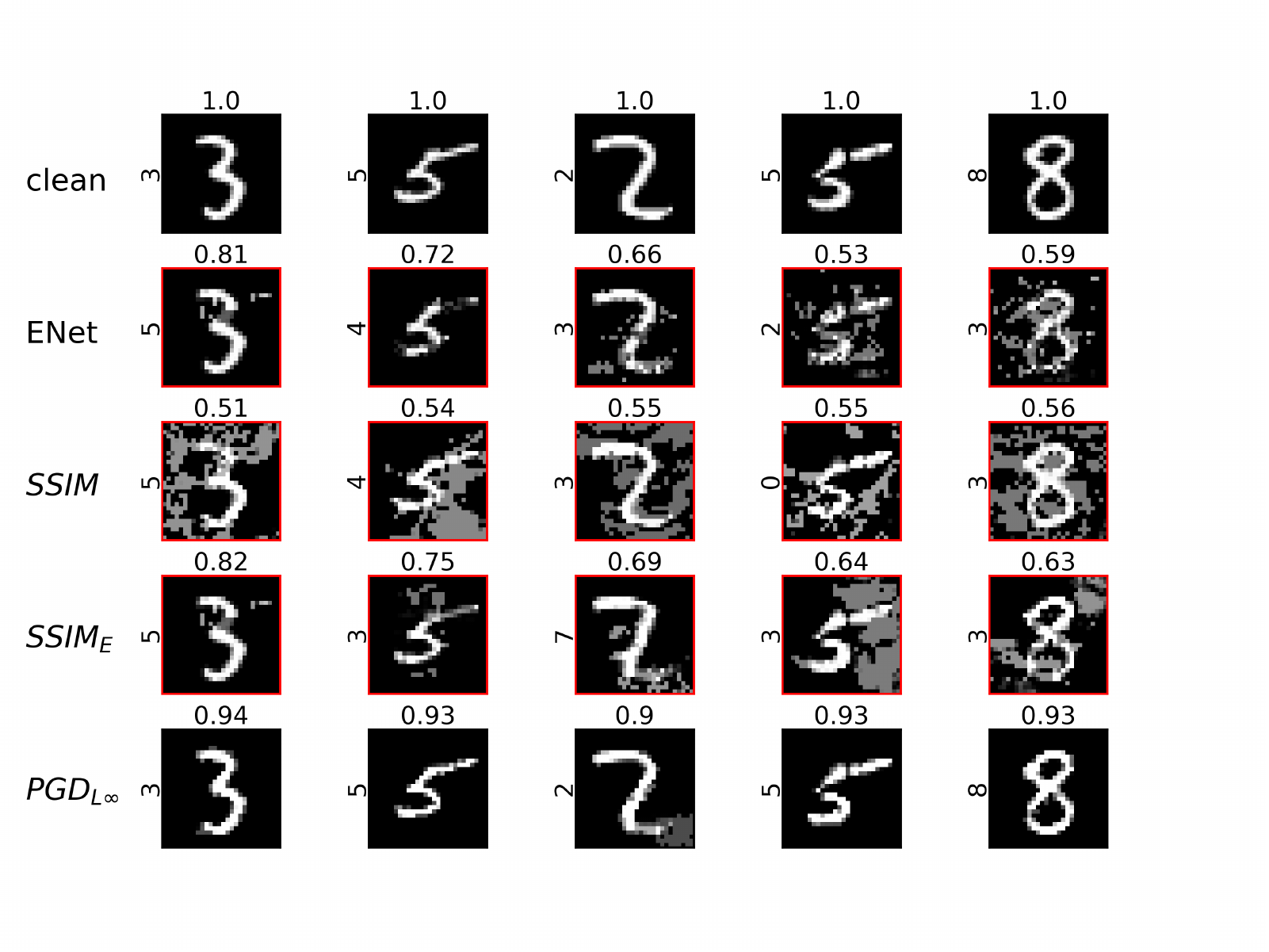} \\[-1em]
    (a) $\ENET$ &  (b) $\SSIM$
    \end{tabular}
\caption{Adversarial images for the $\ENET$ and the $\SSIM$ attacks with smallest SSIM values on the MNIST dataset.}
    \label{fig: mnist_adv_low1}

\end{figure}

The $\PGD$ attack was tested with the same parameters as the ones used for generating adversarial images during adversarial training. It can be seen that the adversarial training indeed helped, and the attack succeeded on the test data only in 7\% of the images. On the other hand, the other three attacks, $\ENET$, $\SSIM$ and $\SSIME$ achieve 100\% (or almost 100\% for $\SSIM$) success rate.
We analyze the quality of these attacks using the SSIM values of the adversarial images. First we use SSIM to find the most distorted images; this is shown in \Cref{fig: mnist_adv_low1} for the $\ENET$ and $\SSIM$ attacks.
The figure shows the images for which $\ENET$, resp. $\SSIM$, produce adversarial images with the lowest SSIM values, together with the adversarial attacks generated by the other methods for the same images. The SSIM values and the perceived labels are shown for each image (on top and on the left hand side, resp.), and red frames indicate if an attack is successful, that is, the resulting image is misclassified.

It can be seen that the adversarial images generated using the $\ENET$ attack with the smallest SSIM values are destroyed to an extent that they have become unrecognizable. On the other hand, for the same images, the $\SSIM$ attack generates successful adversarial images which have relatively higher SSIM values and can be easily recognized by a human oracle. $\SSIME$ also improves quite a lot over the images generated by $\ENET$.
Taking a closer look at digit 3 in  \Cref{fig: mnist_adv_low1} (a), shown in \Cref{fig: need_for_ssim} together with $L_p$-distortion and SSIM values, one can see the positive effect of $\SSIME$ over $\ENET$, as the images produced by the two attacks have almost identical $L_1,L_2$, and $L_\infty$ distances from the original, but the SSIM value of the image produced by $\SSIME$ is much higher and the digit is easy to recognize for a human oracle, unlike the one produced by the $\ENET$ attack.
These (and several other similar) images) raise serious concerns not only about the 100\% success rate of the $\ENET$ attack or other adversarial attacks proposed in literature, and support the observations of \citet{sharif2018suitability} about the unsuitability of $L_p$-distances to asses the quality of adversarial images. Note that here the $\SSIME$ attack produces images with better perceptual quality compared to both $\ENET$ and $\SSIM$ attacks, while the $\PGD$ attack is unsuccessful on these, supposedly hard-to-perturb, images.

\begin{figure}[t]
\centering
\begin{tabular}{cc}
\includegraphics[height = 6.5cm, trim={0.3cm 0cm 1.2cm 0.1cm}, clip]{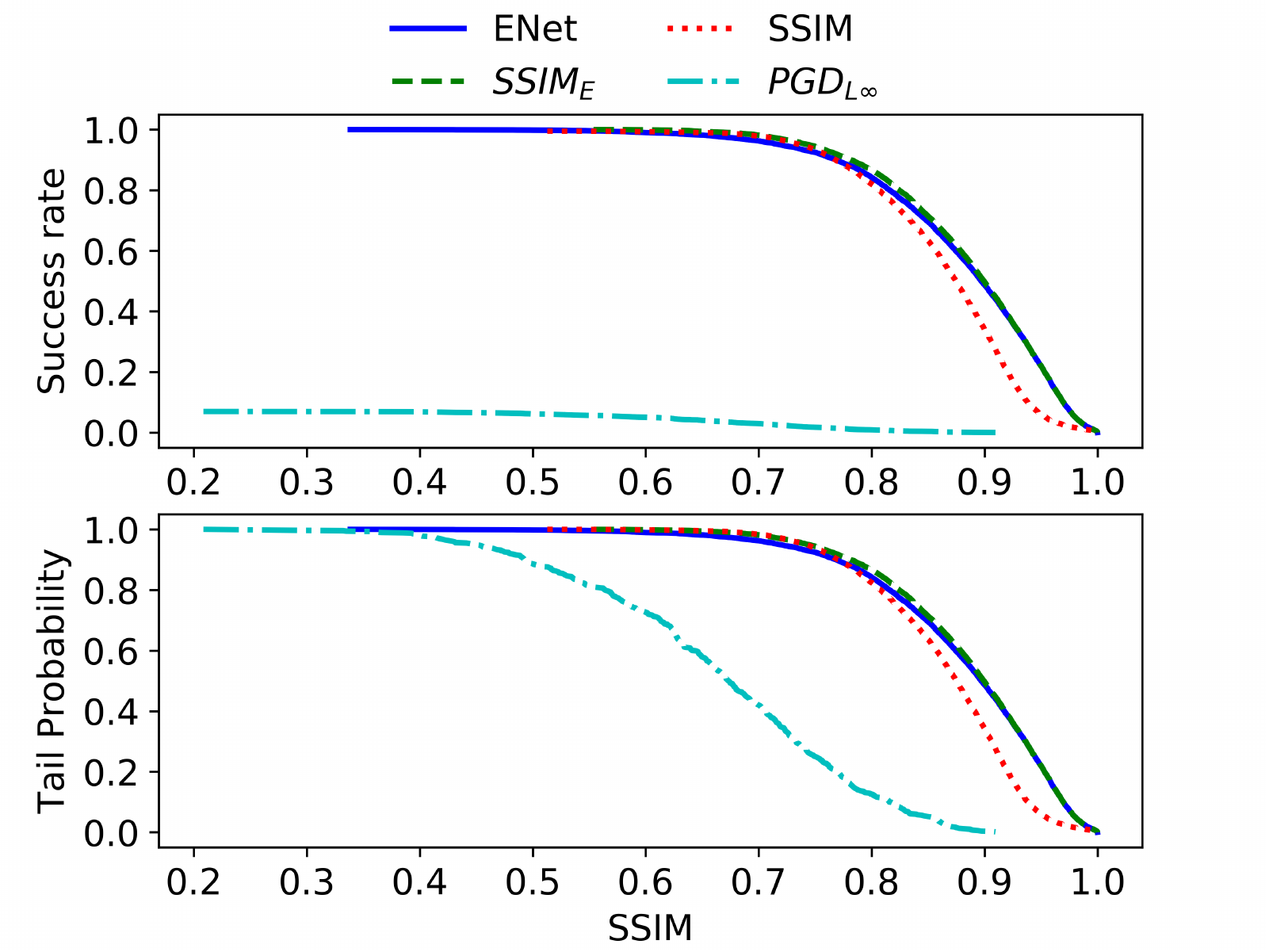} &     \includegraphics[height = 6cm, trim={0.3cm 0cm 2cm 1.5cm}, clip]{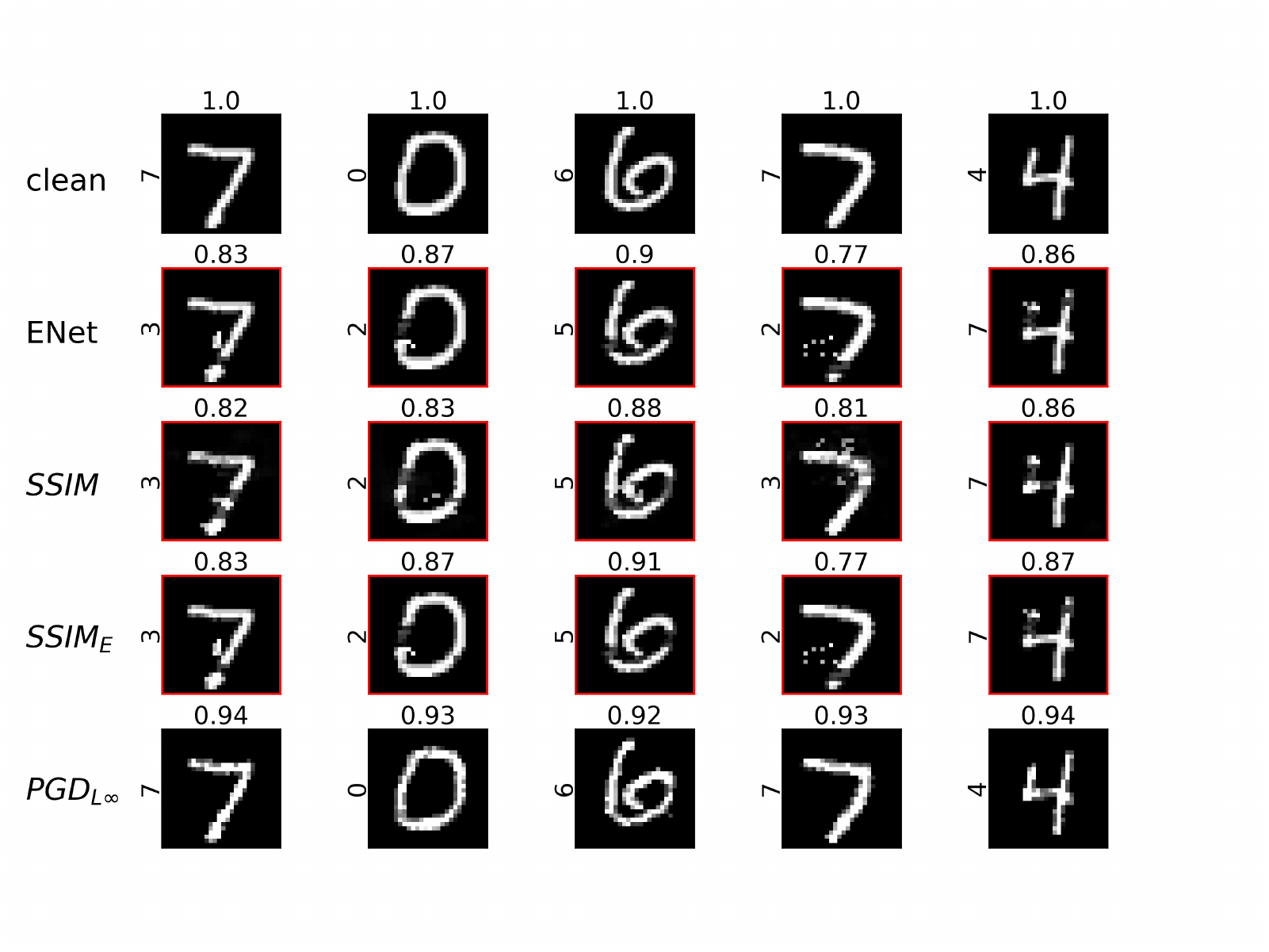} \\
    (a) &

     (b) \\
    \end{tabular}
\caption{(a) Success rate and proportion of adversarial images as a function of the minimum required SSIM value for adversarial attacks on the MNIST dataset. (b) Adversarial images for the $\SSIM$ attack with SSIM values of at least 0.8.}
    \label{fig: mnist_adv_ssim_0_8}
\end{figure}

In order to further evaluate the performance of different adversarial attacks in terms of the SSIM values of successful adversarial examples, we show the success rate of different adversarial attacks as a function of SSIM values in \Cref{fig: mnist_adv_ssim_0_8} (a).  
It can be seen that for the proposed $\SSIM$ and $\SSIME$ attacks, even the worst case SSIM value is above 0.5, which is significantly higher than both for the $\ENET$ and $\PGD$ attacks. In the bottom part of figure (a), we show for each attack the proportion of the adversarial images above a certain SSIM value (the \emph{tail probability}). As almost all adversarial images generated by $\ENET$, $\SSIM$, and $\SSIME$ are successful, the success rate at a given SSIM level is approximately the same as the tail probability. 
The three attacks are equally good when the requirement is to achieve a minimum SSIM value up to 0.8, which guarantees high-quality images, as shown in \Cref{fig: mnist_adv_ssim_0_8} (b); notably, the success rate of the attacks requiring this minimum SSIM value is above 80\%. For higher minimum SSIM values, $\ENET$ becomes somewhat better than $\SSIM$, although $\SSIME$ remains the best.The produced images are also of high quality, as demonstrated in \Cref{fig: mnist_adv_ssim_0_8} (b), showing some images with SSIM around and above 0.8, where the attacks achieve over 80\% success rate.
On the other hand, $\PGD$ is much less effective, e..g., the 30\% of samples distorted below an SSIM value of 0.6 only result in a few percent success in generating adversarial images. $\PGD$ also produces images with low SSIM value in general (e.g., the highest value achieved is approximately 0.9).

\subsection{Adversarial Attacks on the CIFAR-10 Dataset} 
 
In this section we consider three different classifiers for CIFAR-10, trained to be robust against adversarial attacks:
(i) $\Madry$ defense: a Wide-Resnet 32-10-based model of \citet{madry2017towards} 
trained by adversarial training with 7-step $\PGD$ attack with $\epsilon = \frac{8}{255}$.\footnote{We use the pretrained model from the repository provided by \citet{madry2017towards}.} (ii) Free adversarial training: a recently proposed training scheme by \citet{shafahi2019adversarial} which performs the adversarial training without incurring the extra cost used for generating adversarial images by \citet{madry2017towards}, using the same Wide-Resnet 32-10 as $\Madry$ defense; we refer to this defense scheme as $\Free$ defense\footnote{We train a robust model using the code provided with the paper \cite{shafahi2019adversarial}.}. (iii) Feature scattering: the recent feature-scattering-based adversarially trained model of \cite{feature_scatter} using a Wide-ResNet 28-10 model of \citet{BMVC2016_87}, referred to as $\Feature$ defense\footnote{We use the pretrained model from the repository provided by \citet{feature_scatter}.}. 

All three of these defense schemes have been shown to be robust against the 20-step $\PGD$ attack when the attack uses either the standard cross-entropy-loss ($\xent$) or the Carlini-Wagner loss ($\CW$) \citep{carlini2017towards} (given in Eq.~\ref{eq: car_loss} with $c=1$ and $\K=0$) in the optimization to find an adversarial example. 
We then consider the performance of these adversarially trained networks when the perturbation $\epsilon$ used in a 20-step $\PGD$ attack increases beyond the standard $\frac{8}{255}$ used in evaluating these defense schemes in the literature. Analyzing the quality of the produced adversarial images, we first demonstrate (similarly to our experiments on MNIST) that using $L_p$ norms to quantify the quality of the adversarial images can be quite misleading, giving hard-to-interpret accuracy results when the perceptual quality of these adversarial examples are not considered.  The classification accuracy on the adversarially perturbed test sets with different perturbation limits $\epsilon$ (including the clean test set accuracy with $\epsilon=0$) is shown in \Cref{tab: classification_accuracies} (the classifiers are denoted by the name of the model, $\Madry$, $\Free$, or $\Feature$, with a subscript referring to the loss).

\begin{table}
\begin{center}
\begin{small}
\begin{sc}
\resizebox{0.6\columnwidth}{!}{%
\begin{tabular}{lccccc}
\toprule
Defense & Clean & $\epsilon = \frac{8}{255}$ & $\epsilon = \frac{12}{255}$ & $\epsilon = \frac{16}{255}$ & $\epsilon = \frac{20}{255}$ \\
\midrule

$\Madryx$    & 87.25 & 45.86 & 28.68 & 19.69 & 14.46 \\
$\Madryc$    & 87.25 & 47.01 & 30.22 & 20.41 & 14.53 \\
$\Freex$     & 86.05 & 47.04 & 27.52 & 15.63 & 8.39 \\
$\Freec$     & 86.05 & 47.23 & 28.31 & 16.24 & 8.7 \\
$\Featurex$  & 89.98 & 70.81 & 67.52 & 64.64 & 61.55 \\
$\Featurec$  & 89.98 & 59.63 & 54.53 & 50.62 & 47.01 \\
\bottomrule
\end{tabular}
}
\end{sc}
\end{small}
\end{center}
\vskip -0.2in
\caption{Classification accuracy (in \%) for $\PGD$ attack with different maximum perturbations.} 
\label{tab: classification_accuracies}
\end{table}

\Cref{tab: classification_accuracies} shows that both $\Madry$ defense and $\Free$ defense break down as soon as the perturbation size is increased beyond $\epsilon = \frac{8}{255}$, which is used in the training of these networks. $\Feature$ defense is the most robust among our models, achieving non-trivial accuracy even for large values of $\epsilon$.

\begin{figure}[t]
     \centering
    \includegraphics[width=0.53\columnwidth, trim={0cm 0.2cm 0.2cm 0.0cm},clip]{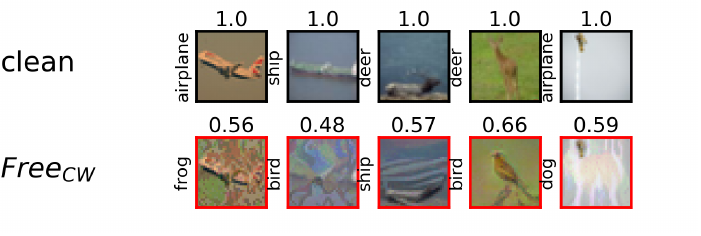} 
    \caption{CIFAR-10 dataset: Low-quality adversarial images for the $\PGD$ attack and $\Free$ defense.}
    \label{fig: cifar10_adv_20_unfiltered_1_free_cw}
\end{figure}

However, inspecting the actual adversarial images used in the attacks, it can be observed that in many cases the image has been either destroyed beyond recognition or modified to represent another class. As an illustration, \Cref{fig: cifar10_adv_20_unfiltered_1_free_cw} 
shows images with low SSIM values (from all the attacks) for $\epsilon= \frac{20}{255}$ (images for more defense schemes are shown in \Cref{fig: cifar10_adv_20_unfiltered_1}(a) in the appendix).
For example, it can be seen that for the $\Free$ defense with $\CW$ attack, clean images of airplane and deer now indeed contain images of a frog, a dog and a bird.
Hence, the reduced classification accuracy at larger perturbations for the $\PGD$ attack  
does not represent the true robustness of these networks (interestingly, as demonstrated in \Cref{fig: cifar10_adv_20_unfiltered_1}(a), $\Feature$ defense maintains higher accuracy for these modified images with larger perturbations than $\Free$ or $\Madry$).

To improve the perceptual quality of the 20-step $\PGD$ attack, we can select a successful adversarial image from all the steps with the highest SSIM value, which we call the \emph{SSIM-filtered} attack (this can be applied to other iterative attacks).
A higher SSIM value typically provides an image with better quality, as demonstrated by randomly selected examples for the $\Feature$ defense with $\PGD$ attack and $\CW$-loss in \Cref{fig: cifar10_adv_20_ssim_scale} in the appendix. These images also indicate (which can be observed more generally) that an SSIM value of about 0.7 is sufficient to have recognizable image classes. \Cref{fig: cifar10_adv_20_unfiltered_1}(b) in the appendix
shows adversarial images with the smallest SSIM values above 0.7; comparing with \Cref{fig: cifar10_adv_20_unfiltered_1}(a) 
demonstrates that this simple SSIM-filtering can significantly improve the perceptual quality of the resulting adversarial images.

\begin{figure}[t]
\centering
      \subfloat[$\Feature$ defense]{\includegraphics[width=0.5\textwidth]{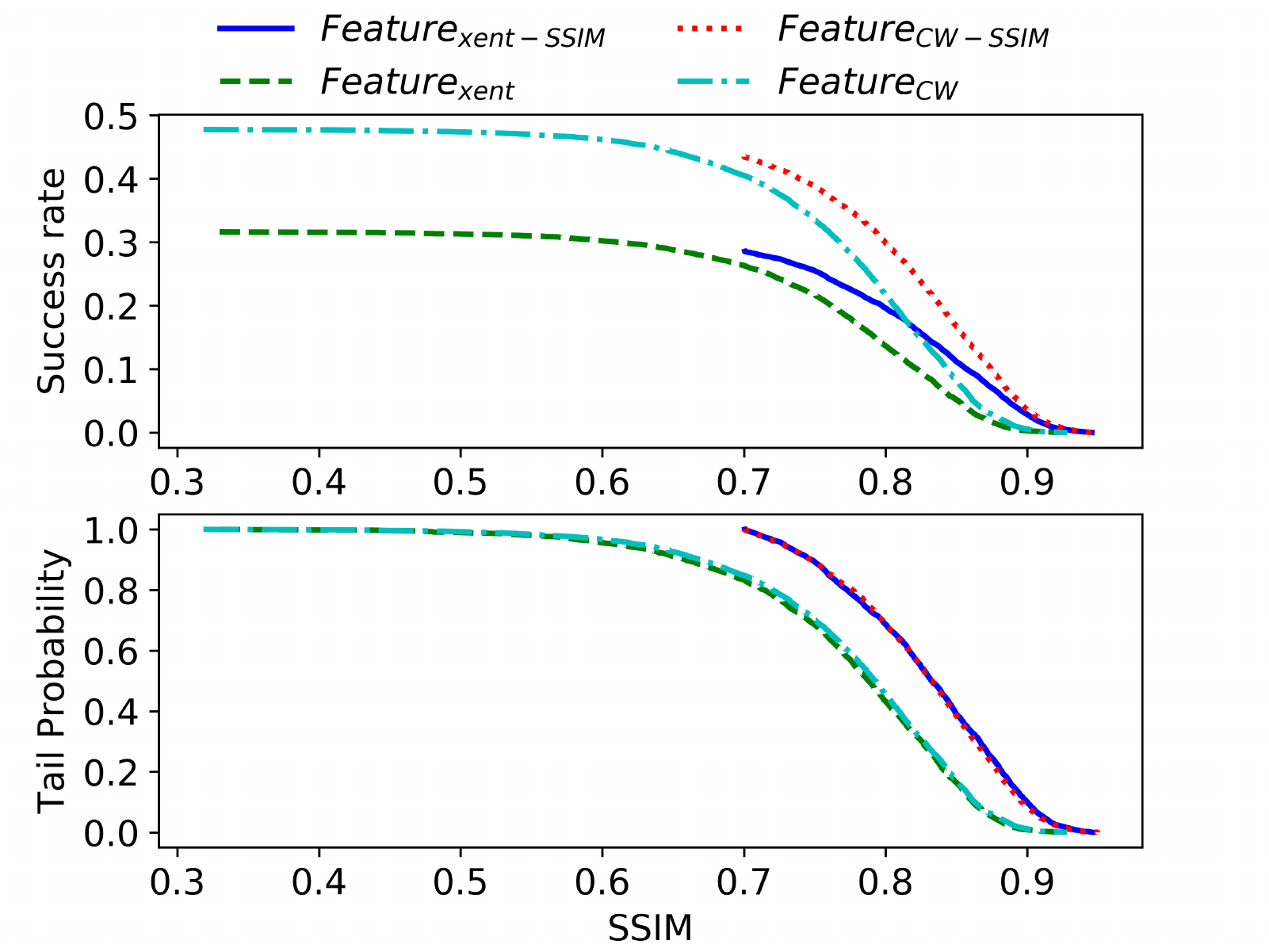}}
      \subfloat[$\Free$ defense]{\includegraphics[width=0.5\textwidth]{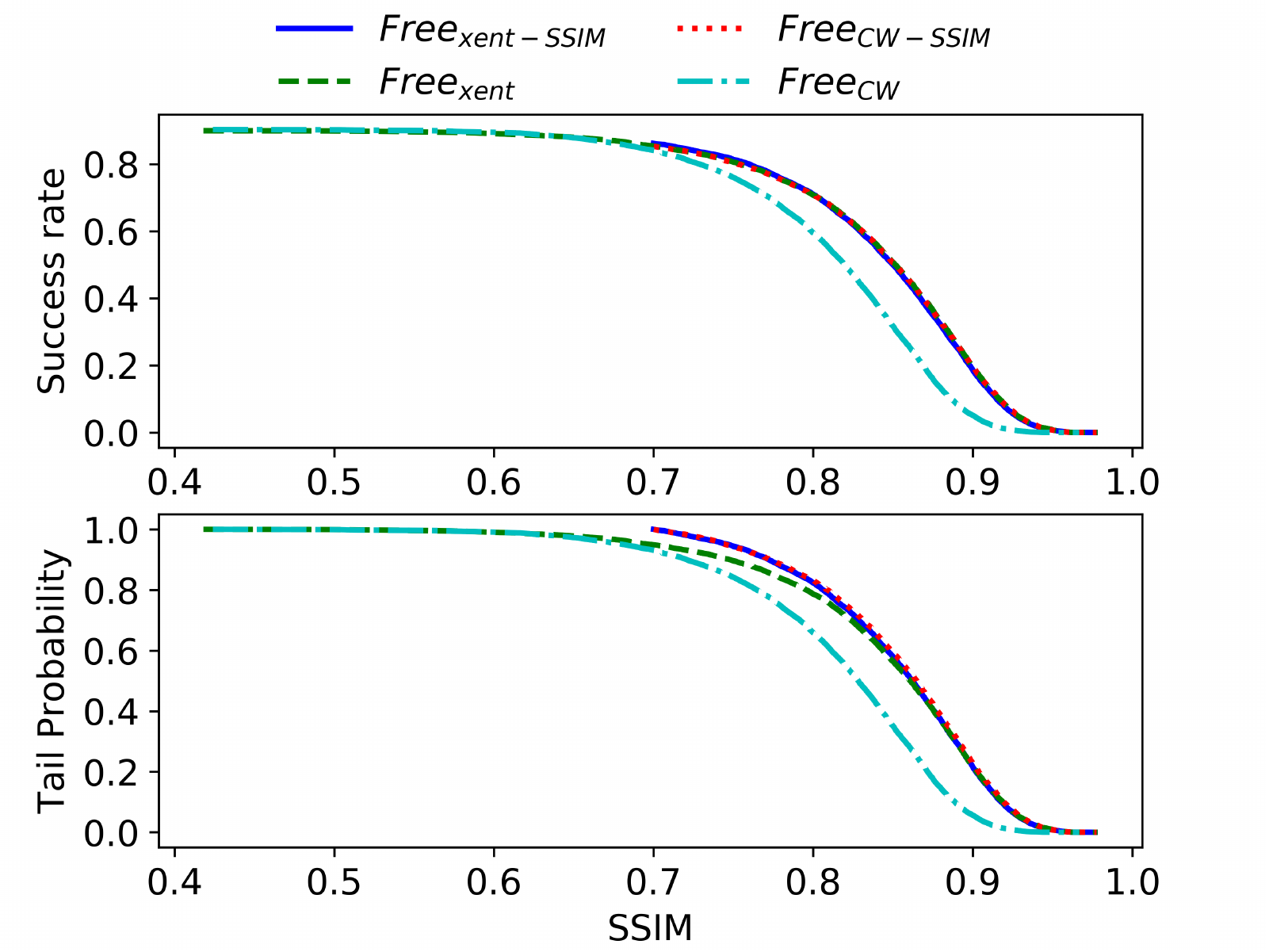}}\\
      \subfloat[$\Madry$ defense]{\includegraphics[width=0.5\textwidth]{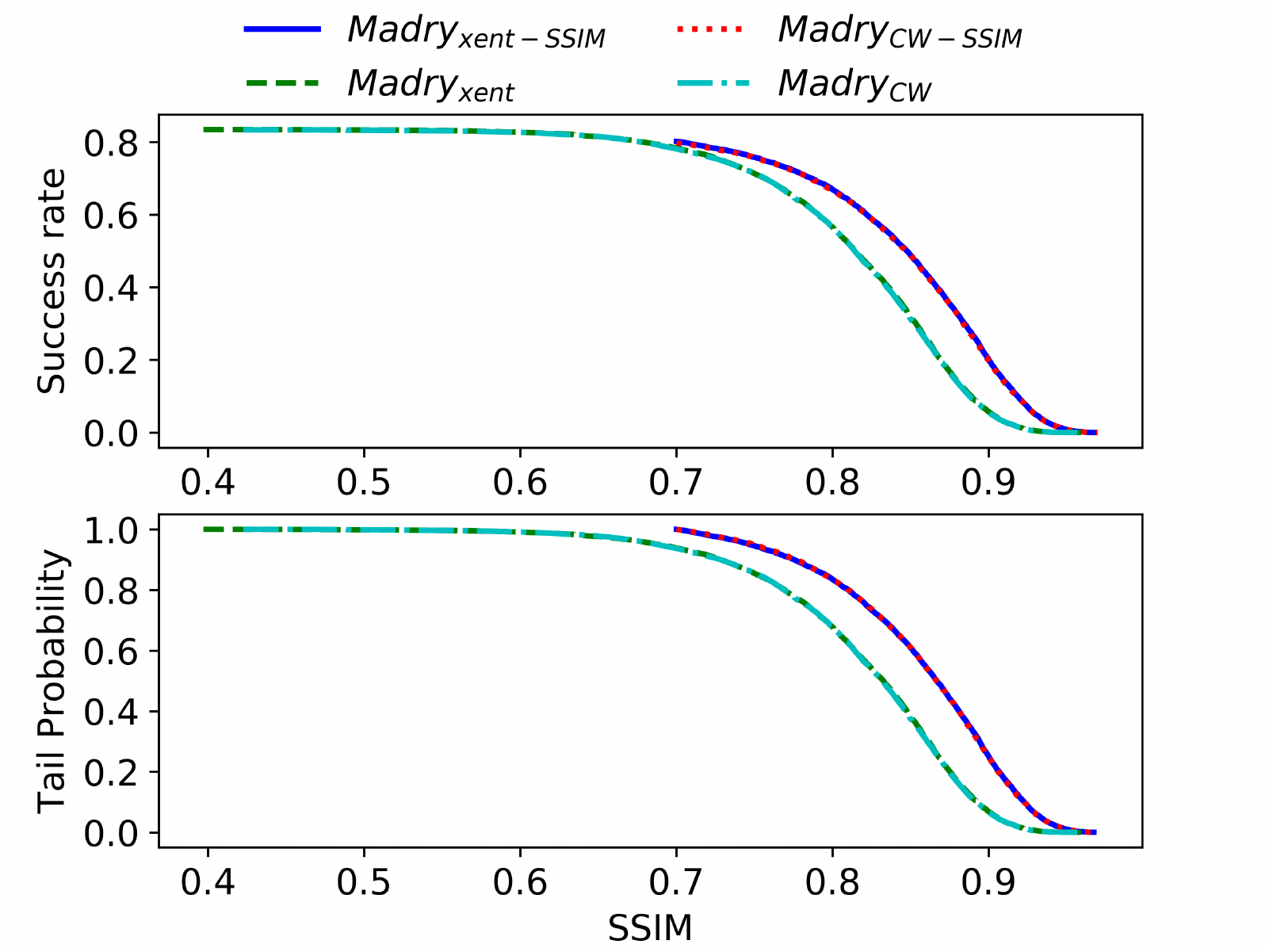}}
      \caption{CIFAR-10 dataset: Attack success rate comparison for $\PGD$-attack and SSIM-filtered $\PGD$-attack.}
      \label{fig: cifar10_success_SSIMf_vs_ssimuf}
\end{figure}

\Cref{fig: cifar10_success_SSIMf_vs_ssimuf} shows the success rate of this SSIM-filtered $\PGD$ attack against the different defense schemes (trained with $\xent$ and $\CW$ loss) as a function of the minimum required SSIM value, compared to the original attacks where the SSIM value of the last step is used (the proportion of the adversarial images with at least the given SSIM value  
is also shown). It can be seen that using SSIM-filtering in the attack results in adversarial images with improved SSIM values, for example,  
when the SSIM of a successful adversarial example is constrained to be at least 0.8, there is about a 6-11\% increase in the attack success rate for all defense schemes for the SSIM-filtered attack. 
Furthermore, comparing the results to \Cref{tab: classification_accuracies}, one can observe that the attack successes for an SSIM constraint of at least 0.7 are only slightly deteriorated and almost achieve the same success rate as if we do not filter for the SSIM values. The graphs also demonstrate that the $\PGD$ attack can achieve high success rate against the considered defense schemes even if we require relatively high quality adversarial images.

\begin{figure}[t]
\centering
      \subfloat[Accuracy]{\includegraphics[width=0.5\columnwidth, trim={5.5inch 13.6inch 4.35inch 16.8inch},clip]{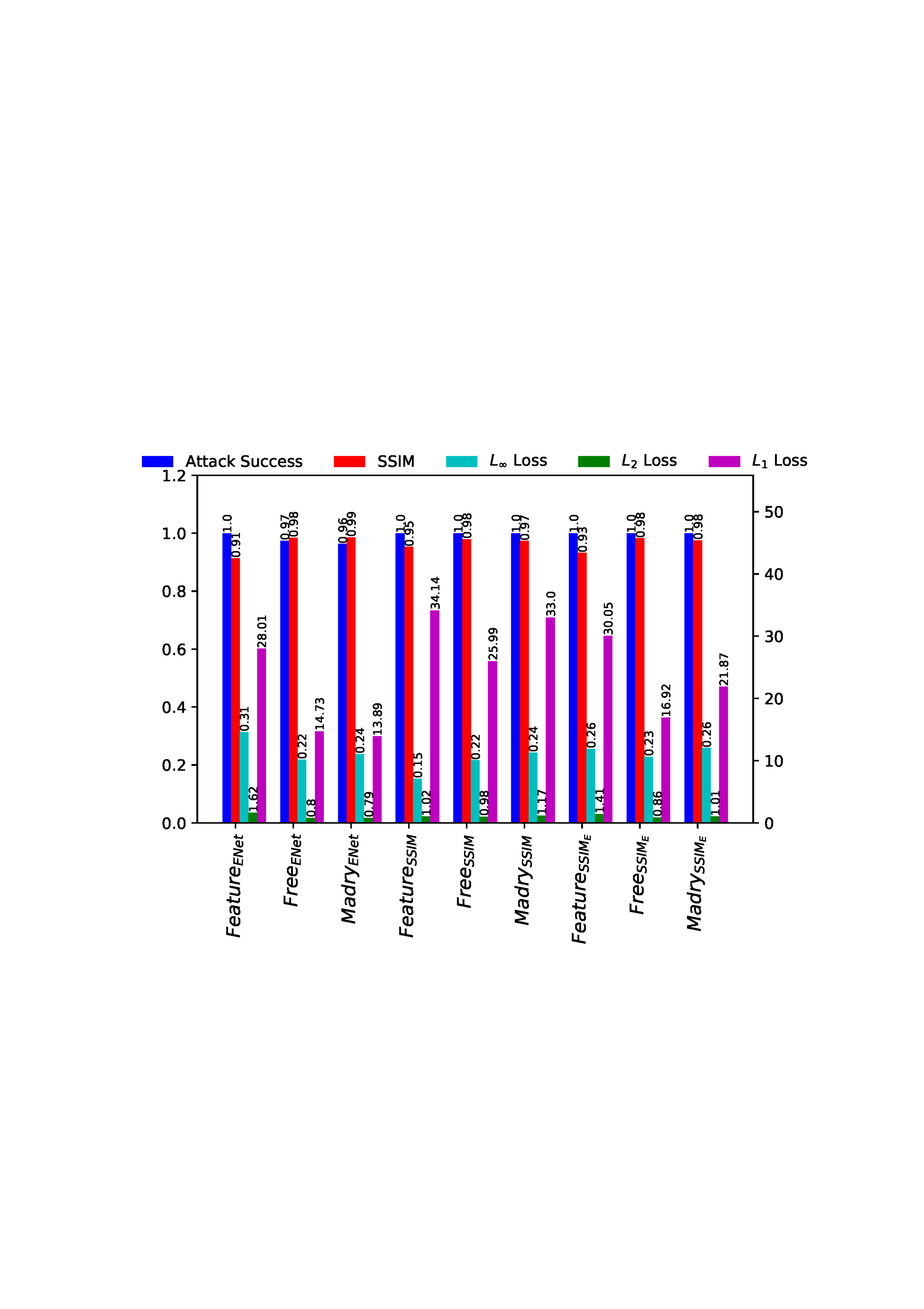}}
      \subfloat[Success rate]{\includegraphics[height = 6.2cm, trim={0.4cm 0.2cm 1.6cm 1.4cm}, clip]{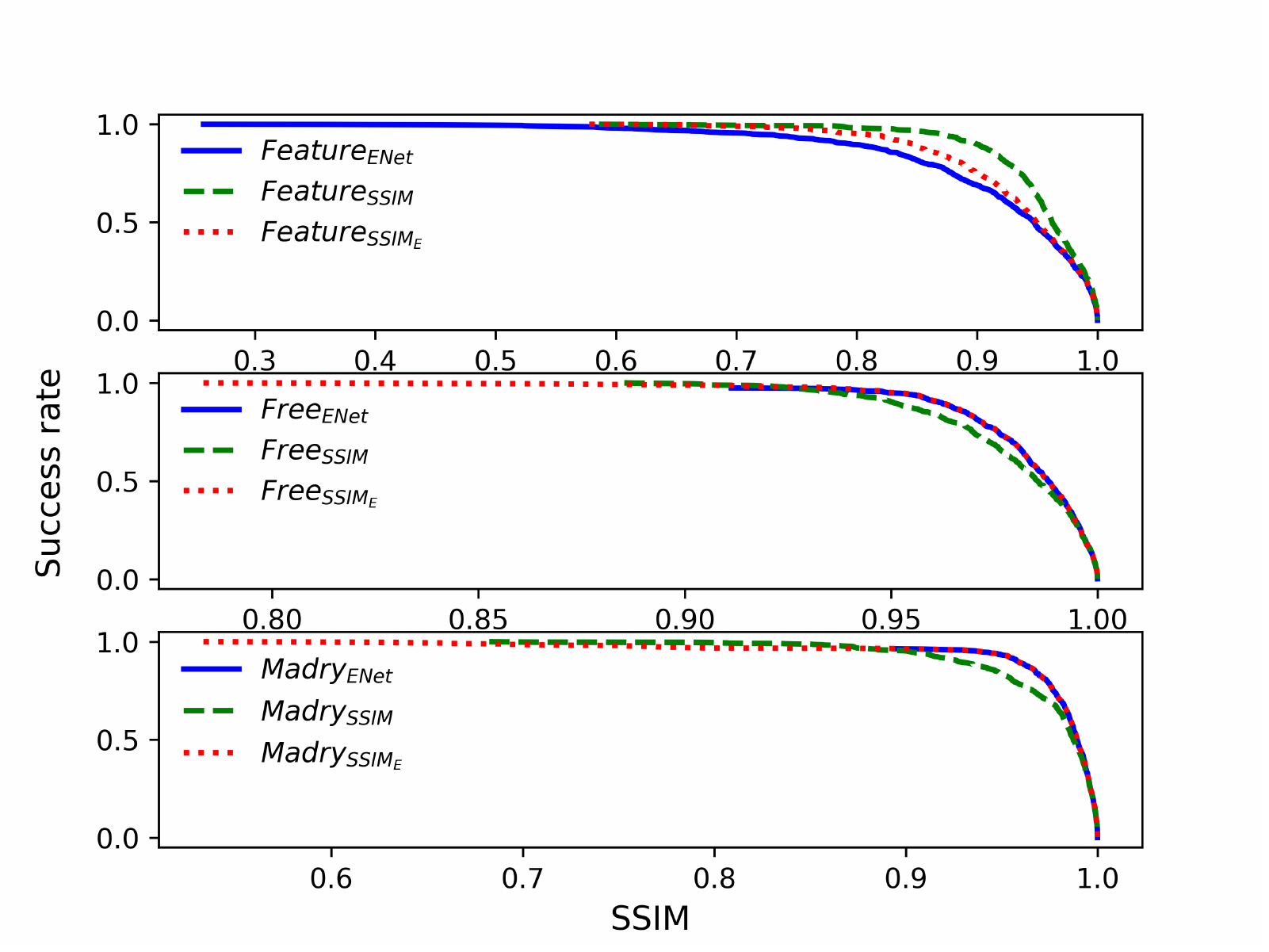}} 
      \caption{CIFAR-10 dataset: Accuracy and attack success rate comparison for the $\ENET$,  $\SSIM,$ and $\SSIM_{E}$ attacks.}
      \label{fig: cifar10_success_ENet_vs_SSIM}
\end{figure}

Next we evaluate the robustness of the $\Madry$, $\Free$, and $\Feature$ defense schemes against our stronger attacks, $\ENET$, $\SSIM$, and $\SSIME$, which produce adversarial perturbations not limited by any $L_{p}$-norm. $\ENET$ is used with the same parameters as for MNIST. For $\SSIM$ and $\SSIME$, all parameters are the same as for MNIST, except for the thresholds $\zeta_1$ and $\zeta_2$, which are chosen to be 0.99, as the SSIM values are relatively higher for CIFAR-10 even for larger perturbations (this is due to the averaging of SSIM over the different color channels).\footnote{To save on computation (as the networks are quite large), these experiments were run on 1000 images from the CIFAR-10 test set.} The achieved accuracy, average distortions, and success rate are shown in \Cref{fig: cifar10_success_ENet_vs_SSIM} (in the figure, the name of the attack method is added as a subscript to the name of the defense scheme). We can see that the SSIM values on average are very high for all attacks. On the other hand, the successful attacks sometimes introduce some artifacts in the images, which are clearly visible (see \Cref{fig: cifar10_adv_soa_ENET_lowest_1} in the appendix), although typically do not interfere with the class of the new images. It can be seen that for $\Feature$ defense (the strongest defense against the $\PGD$ attack), the $\ENET$ attack results in a large number of adversarial images having SSIM below 0.7, which accounts for 4.5\% of all successful adversarial images. In contrast, for $\SSIM$ only 0.65\% of the successful adversarial images have SSIM below 0.7. 
Perhaps surprisingly, the $\ENET$ attack is unable to achieve 100\% adversarial success rate against the $\Free$ and $\Madry$ defenses, but the $\SSIM$ and $\SSIME$ attacks achieve 100\% adversarial success rate against all defense schemes when no constraint is imposed on the SSIM value. 
One can also see that the $\SSIM$ attack (and also $\SSIME$) provides a significant improvement in the SSIM values of the successful adversarial examples compared to $\ENET$, especially for the most effective defense scheme, $\Feature$. For example, at an SSIM level of 0.8, the $\SSIM$ attack achieves a 10\% higher success rate than $\ENET$, which goes up to 20\% when the SSIM is 0.9 (which accounts for an almost 30\% relative improvement).
\Cref{fig: cifar10_adv_soa_ENET_lowest_1} in the appendix 
shows the successful adversarial images for the $\ENET$ attack with the lowest SSIM values for the $\Feature$ defense, which are nearly destroyed, while the corresponding adversarial images obtained by $\SSIM$ and $\SSIME$ have high SSIM values and are perceptually very similar to the original images. For a fair comparison, \Cref{fig: cifar10_adv_soa_SSIM_lowest_1} in the appendix shows the successful adversarial images for the $\SSIM$ attack with the lowest SSIM values for the $\Feature$ defense, which are not destroyed beyond recognition. Visually inspecting the adversarial images from the attacks for different SSIM values of at least 0.7, as shown  in \Cref{fig: cifar10_adv_images_ENet_SSIM_scale} in  the appendix 
, it can be seen that even for very high values of SSIM (at least 0.95) modifications are not truly imperceptible.

Hence, it can be concluded that even though the state-of-the-art $\Madry$, $\Free$ and $\Feature$ defense schemes achieve robustness against bounded adversarial perturbations, their performance deteriorates quickly as soon as the size of the perturbations increase even for the $\PGD$ attack. In order to keep these potentially unbounded perturbations perceptually similar to the original images, we can use SSIM to disallow images of too low quality. Allowing only images with an SSIM value of at least 0.7,  the attacks achieve similar performance as the unbounded attacks but with better perceptual quality. Compared to the $\Madry$ and $\Free$ defense schemes, $\Feature$ defense is the most robust against these unbounded perturbations produced using the $\PGD$ attack, but the more advanced optimization-based attacks ($\ENET$, $\SSIM$ and $\SSIME$) also break this defense completely even though the successful adversarial images, especially for $\SSIM$ and $\SSIME$, are perceptually very similar to the original ones. This highlights the shortcomings of current measures for the evaluation of robustness of defense schemes against bounded adversarial perturbations, and justifies the use of perception-based attacks, such as our proposed SSIM-based methods.

\section{Conclusions}\label{sec: conclusion}
In this paper, we proposed to use the perceptual similarity measure SSIM as the quality metric instead of standard $L_p$-distances for both adversarial attacks and evaluating the robustness of adversarial defense schemes. Experiments on the MNIST and the CIFAR-10 datasets  demonstrate that our proposed adversarial attacks ($\SSIM$ and $\SSIME$) achieve the same or higher success rate as state-of-the-art attacks using possibly unbounded perturbations (such as the elastic net attack), while producing adversarial images of better perceptual quality. Our experiments also demonstrate that it is possible to completely break recent state-of-the-art defense schemes (such as the feature-scattering defense) with adversarial examples of high perceptual quality.

\bibliographystyle{plainnat}
\bibliography{PCAA}

\clearpage
\newpage
\appendix
\onecolumn
 \begin{center}
    {\bf \large APPENDIX}
  \end{center}

\section{Binary search algorithm to set $c$ in  \eqref{eq: car_loss}}
\label{app:binsearch}

The binary search algorithm to tune $c$ in the loss function  \eqref{eq: car_loss}, as proposed by \citet{chen2017ead} for the $\ENET$ attack, is shown in \Cref{algo: binsearch}. 
In the experiments we ran the search for $K=9$ steps.
\begin{algorithm}[!h]
  \caption{Binary search algorithm for parameter $c$ in \eqref{eq: car_loss}. }
  \label{algo: binsearch}
\begin{algorithmic}
\STATE{\bfseries Input:} $K \in \mathbb{N}$, $c_{init} = 10^{-3}$, classifier $f$, adversarial attack $A$, quality measure $Q $
  \STATE {\bfseries Initialize:} $c = c_{init}$, LowerBound = 0, UpperBound = $10^{10}$ 
  \FOR{k $\in$ K}
      \STATE x + $\delta = A(c,x)$
      \STATE $\hat{s} = f(x + \delta)$, $s = f(x)$
      \IF {$\hat{s} \neq s$}
        \STATE UpperBound = $\min$(UpperBound, $c$)
        \IF{UpperBound $< 10^{9}$}
            \STATE $c = \frac{\text{LowerBound + UpperBound}}{2}$
        \ENDIF
      \ELSE
        \STATE LowerBound = $\max$(LowerBound, $c$)
        \IF {UpperBound $< 10^{9}$}
            \STATE $c = \frac{\text{LowerBound + UpperBound}}{2}$
        \ELSE
            \STATE $c = 10 c$
        \ENDIF
      \ENDIF
  \ENDFOR
  \STATE  Return $x + \delta$ with highest quality measure Q found in the $K$ iterations.
\end{algorithmic}
\end{algorithm}

\newpage
\section{Additional images}
\label{app:ssim_low_feature}

\begin{figure}[!h]
    \centering
    \begin{tabular}{cc}
    \includegraphics[height= 10cm,trim={0cm 0.7cm 0.7cm 0.5cm},clip]{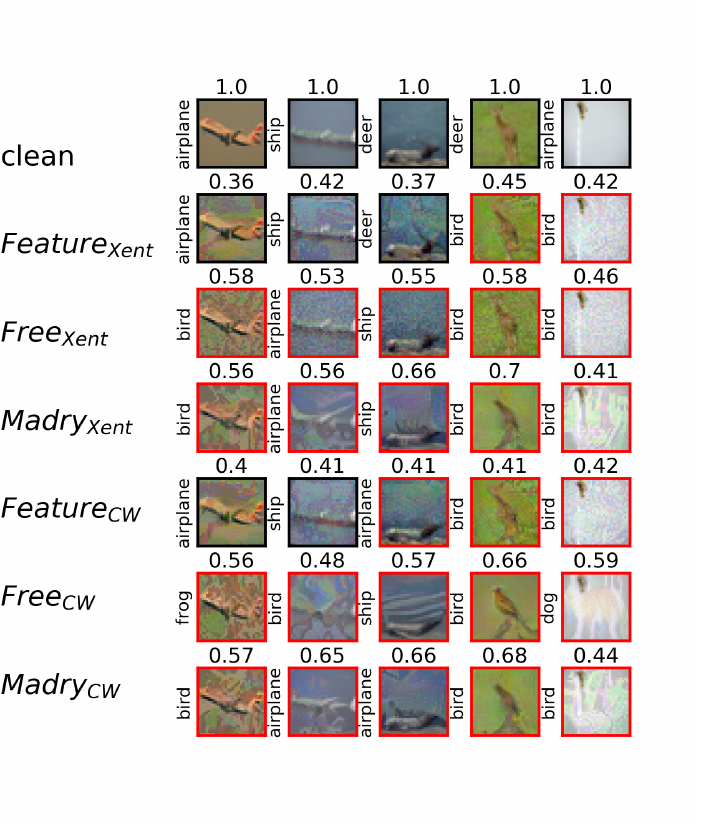} 
    & \includegraphics[height= 10cm,trim={1.80cm 0.7cm 0.5cm 0.5cm},clip]{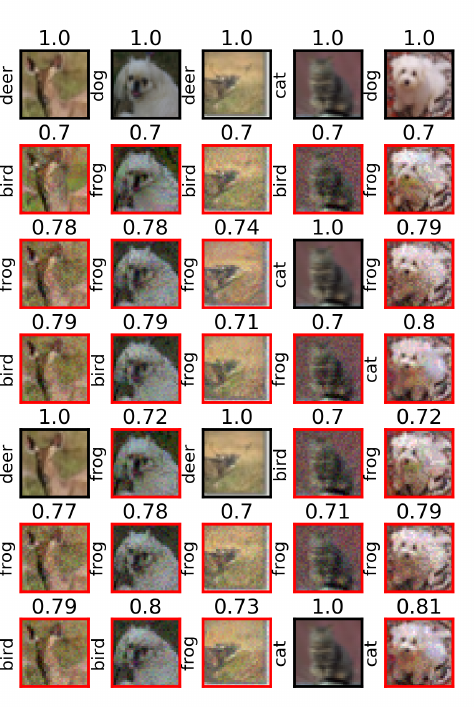} \\
    \small    (a) $\epsilon = \frac{20}{255}$ & \small (b) $\epsilon \leq  \frac{20}{255}$ with SSIM $\geq 0.7$
     \end{tabular}       
    \caption{CIFAR-10 dataset: Adversarial images for $\PGD$ attack.}
    \label{fig: cifar10_adv_20_unfiltered_1}
\end{figure}

\begin{figure}[!h]
\centering
    \begin{tabular}{cc}
     \footnotesize{Clean} & \includegraphics[width=0.90\columnwidth, trim={2.5cm 5.5cm 1cm 5.5cm},clip]{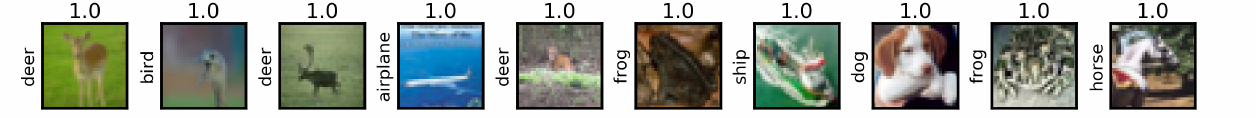}\\
     \footnotesize{$\Featurec$} & \includegraphics[width=0.90\columnwidth, trim={2.5cm 5.5cm 1cm 5.5cm},clip]{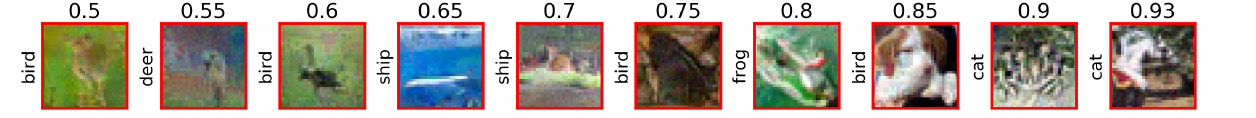}
    \end{tabular}
      \caption{CIFAR-10 dataset: Adversarial images for perturbation $\epsilon = \frac{20}{255}$.}
      \label{fig: cifar10_adv_20_ssim_scale}
\end{figure}

\begin{figure}[t]
    \centering

    \includegraphics[width=1.0\columnwidth,trim={0cm 1cm 1cm 1cm},clip]{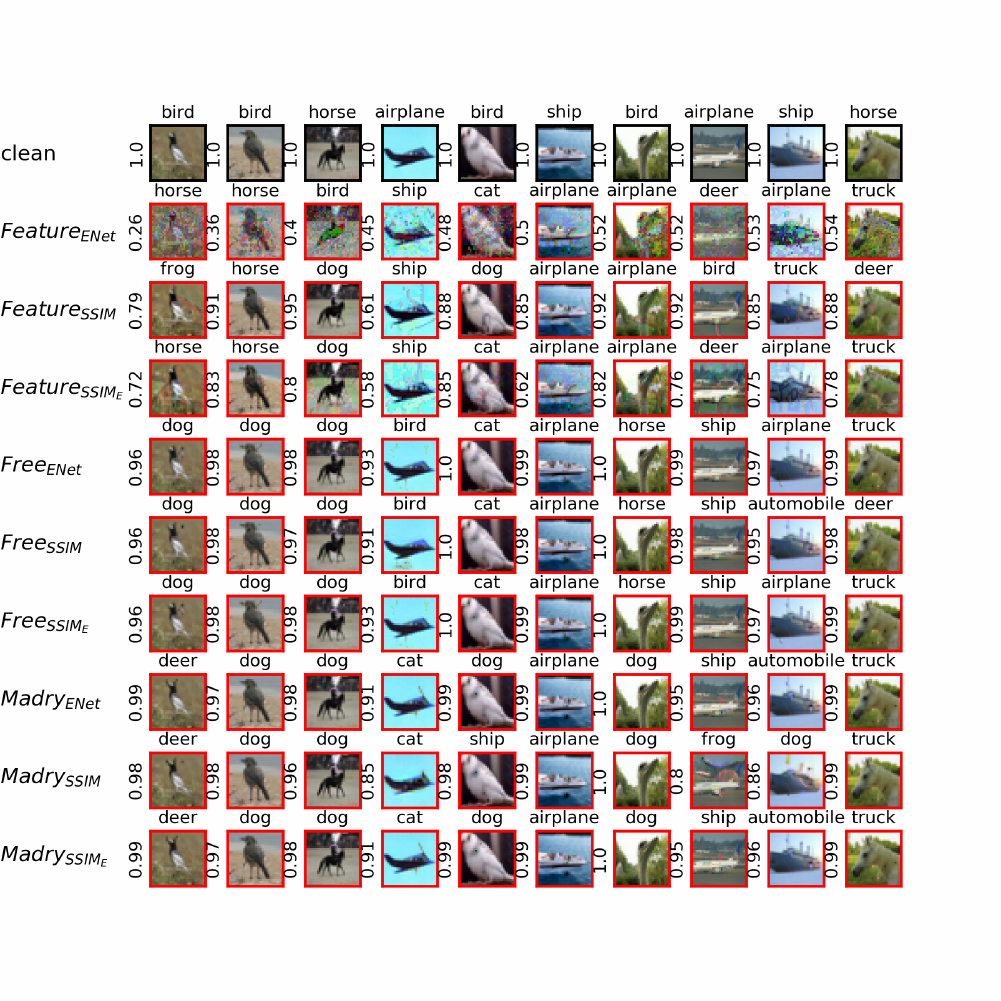} 
    \caption{CIFAR-10 dataset: Adversarial images produced by the $\ENET$ attack with lowest SSIM values.}
    \label{fig: cifar10_adv_soa_ENET_lowest_1}
\end{figure}

\begin{figure}[!h]
    \centering
    \includegraphics[width=1.0\columnwidth,trim={0cm 1cm 0.8cm 1cm},clip]{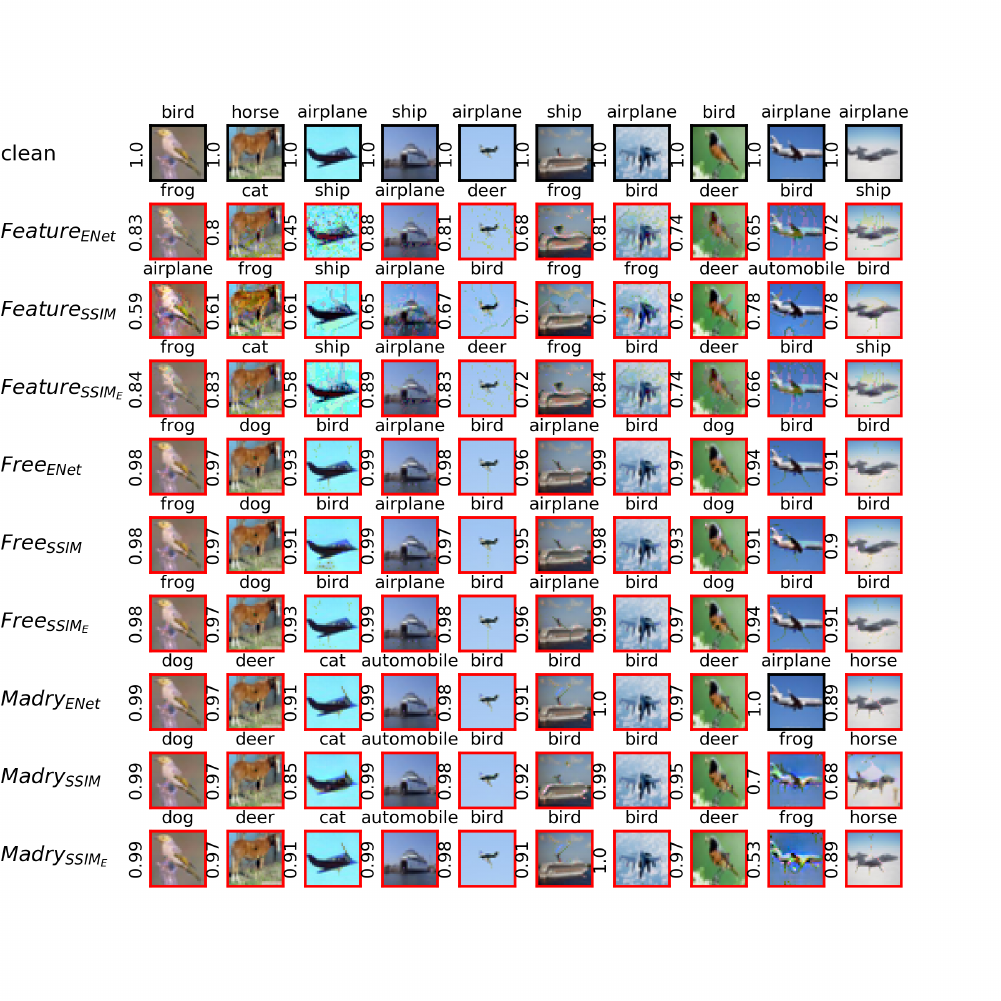} 
    \caption{CIFAR-10 dataset: Adversarial images produced by the $\SSIM$ attack with lowest SSIM values.}
    \label{fig: cifar10_adv_soa_SSIM_lowest_1}
\end{figure}
\begin{figure}[!h]
\centering
    \begin{tabular}{c}
      \subfloat{\includegraphics[width=0.67\columnwidth,trim={0cm 1cm 0.5cm 1cm},clip]{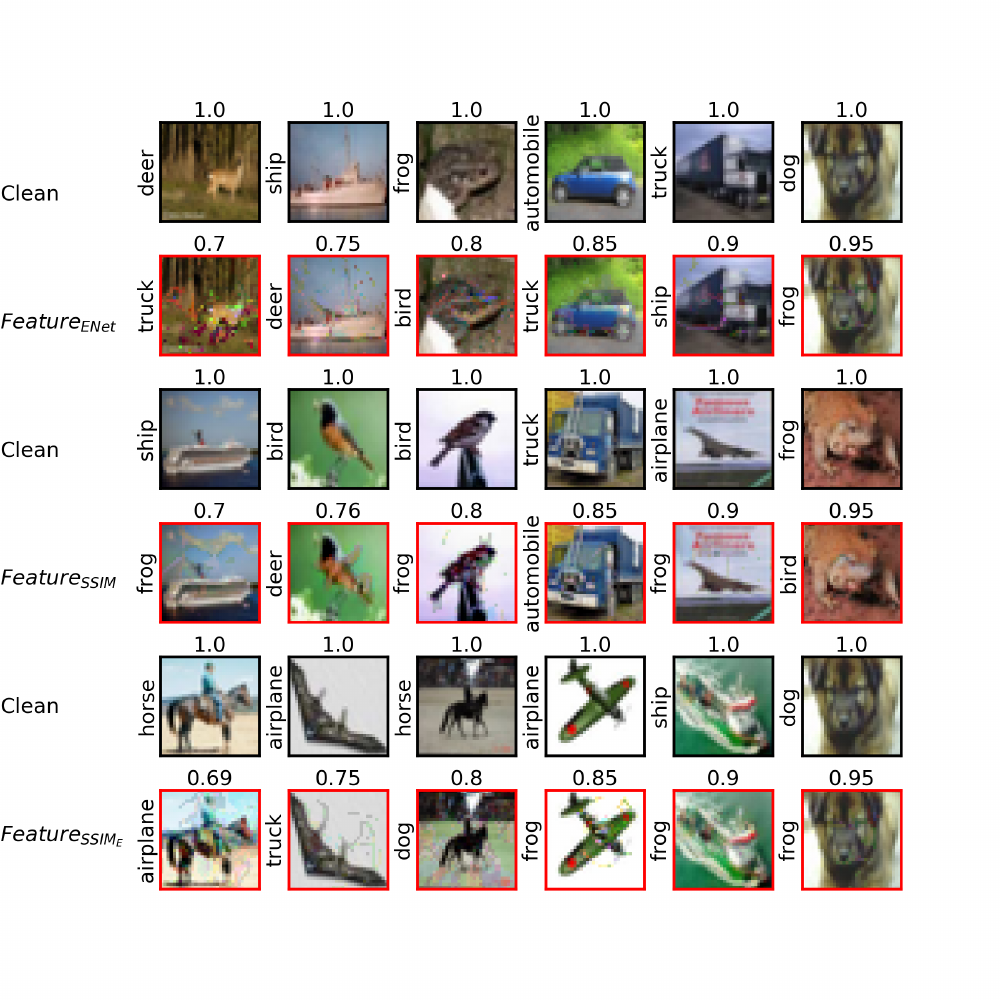}} \\
      (a) \\
       \subfloat{\includegraphics[width=0.67\columnwidth, trim={0cm 1cm 0.5cm 1cm},clip]{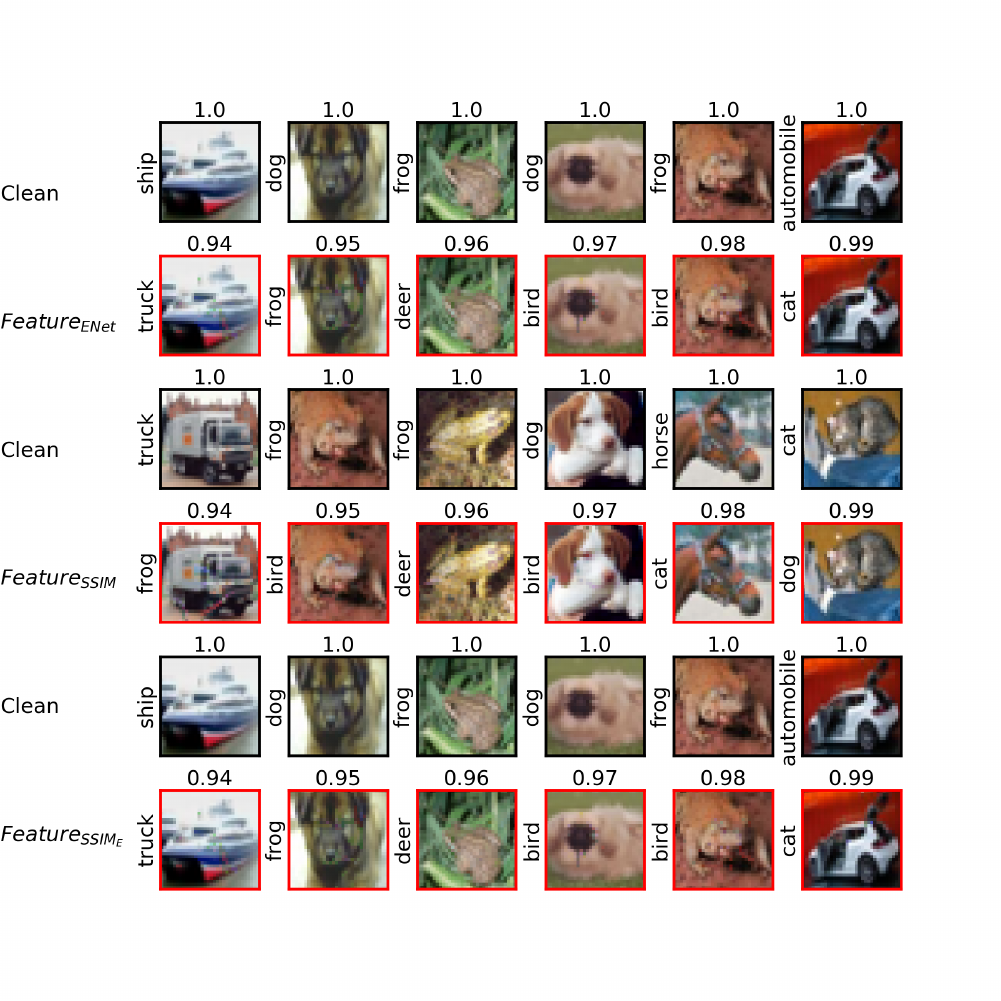}} \\
      (b)

           \end{tabular}
      \caption{CIFAR-10 dataset: Adversarial images for $\ENET$, $\SSIM$ and $\SSIM_{E}$-attack with different SSIM values: (a) low SSIM values; (b) high SSIM values (the attack is shown as the subscript to the name of the defense method).}
      \label{fig: cifar10_adv_images_ENet_SSIM_scale}
\end{figure}

\end{document}